\documentclass[conference,compsoc]{IEEEtran}
\usepackage{cite}
\usepackage{amsmath,amssymb,amsfonts}
\usepackage{algorithmic}
\usepackage{graphicx}
\usepackage{textcomp}
\usepackage{xcolor}
\def\BibTeX{{\rm B\kern-.05em{\sc i\kern-.025em b}\kern-.08em
    T\kern-.1667em\lower.7ex\hbox{E}\kern-.125emX}}

\usepackage{tikz}
\usepackage{dblfloatfix}
\usepackage{multirow}
\usepackage{enumitem}
\usepackage{versions}
\usepackage{xspace}
\usepackage{subfigure}
\usetikzlibrary{shapes,arrows}
\usetikzlibrary{shapes.geometric}
\usetikzlibrary{patterns}
\usetikzlibrary{backgrounds, positioning, fit}
\usepackage{pgfplots}
\usepackage[T1]{fontenc}
\usepackage{multirow}
\usepackage{colortbl}
\usepackage{url}

\DeclareMathAlphabet\mathbfcal{OMS}{cmsy}{b}{n}

\newcommand\topic[1]{{\color{blue}\textbf{#1\\}\xspace}}

\newcommand{\method}{\protect\textsc{GrOVe}\xspace} 

\newcommand{\suspect}{\protect$\mathcal{F}_{?}$\xspace}
\newcommand{\surrogate}{\protect$\mathcal{F}_{s}$\xspace}
\newcommand{\target}{\protect$\mathcal{F}_{t}$\xspace}
\newcommand{\independent}{\protect$\mathcal{F}_{i}$\xspace}
\newcommand{\similarity}{\protect$\mathcal{C}_{sim}$\xspace}
\newcommand{\similaritybold}{\protect$\mathbfcal{C}_{sim}$\xspace}
\newcommand{\classifier}{\protect$\mathcal{C}$\xspace}
\newcommand{\doublesurr}{\protect$\mathcal{F}_{s^2}$\xspace}

\newcommand{\dblp}{\protect\textsc{DBLP}\xspace}
\newcommand{\coauthor}{\protect\textsc{CoAuthor}\xspace}
\newcommand{\citseer}{\protect\textsc{Citeseer}\xspace} 
\newcommand{\acm}{\protect\textsc{ACM}\xspace}
\newcommand{\amazon}{\protect\textsc{Amazon}\xspace}
\newcommand{\pubmed}{\protect\textsc{Pubmed}\xspace}

\newcommand{\targetdata}{\protect$\mathcal{D}_{t}$\xspace}
\newcommand{\surrogatedata}{\protect$\mathcal{D}_{s}$\xspace}
\newcommand{\independentdata}{\protect$\mathcal{D}_{i}$\xspace}
\newcommand{\inputverify}{\protect$\mathcal{D}_{v}$\xspace}
\newcommand{\modelonedata}{\protect$\mathcal{D}_{1}$\xspace}
\newcommand{\modeltwodata}{\protect$\mathcal{D}_{2}$\xspace}

\newcommand{\adv}{\protect{$\mathcal{A}dv$}\xspace}
\newcommand{\verifier}{\protect{$\mathcal{V}er$}\xspace}
\newcommand{\accuser}{\protect{$\mathcal{A}ccuser$}\xspace}
\newcommand{\responder}{\protect{$\mathcal{R}esponder$}\xspace}
\newcommand{\respadv}{\protect{$\mathcal{A}dv.\mathcal{R}$}\xspace}
\newcommand{\accadv}{\protect{$\mathcal{A}dv.\mathcal{A}$}\xspace}

\newcommand{\embds}{\protect{$\mathcal{H}$}\xspace}
\newcommand{\embdsurrogate}{\protect{$h_{s}$}\xspace}
\newcommand{\embdtarget}{\protect{$h_{t}$}\xspace}
\newcommand{\embdindependent}{\protect{$h_{i}$}\xspace}
\newcommand{\embdsetsurrogate}{\protect{$\mathcal{H}_{s}$}\xspace}
\newcommand{\embdsettarget}{\protect{$\mathcal{H}_{t}$}\xspace}
\newcommand{\embdsetindependent}{\protect{$\mathcal{H}_{i}$}\xspace}
\newcommand{\embdsetsuspect}{\protect{$\mathcal{H}_{?}$}\xspace}

\newcommand{\commitment}{\protect{$c$}\xspace}
\newcommand{\committarget}{\protect{$c_t$}\xspace}
\newcommand{\commitsuspect}{\protect{$c_?$}\xspace}
\newcommand{\timestamp}{\protect{$t$}\xspace}
\newcommand{\timetarget}{\protect{$t_t$}\xspace}
\newcommand{\timesuspect}{\protect{$t_?$}\xspace}

\excludeversion{topics}

\includeversion{full}

\excludeversion{short}
\pgfplotsset{compat=1.17}

\usepackage[strings]{underscore}

\begin{document}

\title{\method: Ownership Verification of Graph Neural Networks using Embeddings}

\author{
\IEEEauthorblockN{Asim Waheed}
\IEEEauthorblockA{\textit{University of Waterloo}\\
asim.waheed@uwaterloo.ca}
\and
\IEEEauthorblockN{Vasisht Duddu}
\IEEEauthorblockA{\textit{University of Waterloo}\\
vasisht.duddu@uwaterloo.ca}
\and
\IEEEauthorblockN{N. Asokan}
\IEEEauthorblockA{\textit{University of Waterloo}\\
and Aalto University \\
asokan@acm.org}
}

\maketitle

\begin{abstract}
Graph neural networks (GNNs) have emerged as a state-of-the-art approach to model and draw inferences from large scale graph-structured data in various application settings such as social networking. The primary goal of a GNN is to learn an \emph{embedding} for each graph node in a dataset that encodes both the node features and the local graph structure around the node. 

Prior work has shown that GNNs are prone to  model extraction attacks. Model extraction attacks and defenses have been explored extensively in other non-graph settings. While detecting or preventing model extraction appears to be difficult, deterring them via effective \emph{ownership verification techniques} offer a potential defense. In non-graph settings, \emph{fingerprinting} models, or the data used to build them, have shown to be a promising approach toward ownership verification.

We present \method, a state-of-the-art GNN model fingerprinting scheme that, given a \emph{target} model and a \emph{suspect} model, can reliably determine if the suspect model was trained independently of the target model or if it is a \emph{surrogate} of the target model obtained via model extraction. 
We show that \method can distinguish between surrogate and independent models even when the independent model uses the same training dataset and architecture as the original target model. 

Using six benchmark datasets and three model architectures, we show that \method consistently achieves low false-positive and false-negative rates. We demonstrate that \method is \emph{robust} against known fingerprint evasion techniques while remaining computationally \emph{efficient}. \footnote{To appear in the IEEE Symposium on Security and Privacy, 2024.}
\end{abstract}


\section{Introduction}\label{sec:introduction}

Graph data is ubiquitous and used to model networks such as social networks, chemical compounds, and financial transactions. However, the non-euclidean nature of graph data makes it difficult to analyze using traditional machine-learning algorithms. Unlike Euclidean datasets, where the data points are independent, graph data follows homophily, where similar nodes share an edge. To analyze such graph data, special deep neural networks (DNNs) called Graph Neural Networks (GNNs) have been introduced~\cite{gcn, GraphSAGE, gin, gat}. These models learn an embedding for each node in the graph that encodes the node features and local graph structure. They can perform node classification \cite{gcn, GraphSAGE, gin, gat}, link prediction~\cite{cai2021LinkPrediction, yun2021LinkPrediction, zhang2018LinkPrediction, zhang2021LinkPrediction}, visualization~\cite{wang2021Visualization}, and recommendations~\cite{li2019Recommendation, sun2020Recommendation, wang2019Recommendation2, ying2018Recommendation}. 

Model builders spend significant time and resources to prepare the training data, perform hyperparameter tuning, and optimize the model to achieve state-of-the-art performance. This makes the deployed GNNs a critical intellectual property for model owners.
For instance, Amazon Neptune~\cite{amazonAmazonNeptune} provides a framework to train and use GNNs; Facebook~\cite{facebookGraphML} and Twitter~\cite{twitterGraphML} extensively use graph-based ML models. This broad adoption of GNNs makes them vulnerable to model extraction attacks where an adversary tries to train a local \emph{surrogate} model with similar functionality as the deployed \emph{target} model~\cite{tramer2016ModelExtraction, papernot2017ModelExtraction, okerondy2018ModelExtraction, jagielskiHighFidHighAcc, jagielsiCrypto, buseBoogeyman}.

Model extraction attacks use the input-output pairs from the target model to train the surrogate model. The goal is to achieve similar utility on the primary task for a fraction of the cost. These attacks violate the company's intellectual property and allow further attacks such as evasion using adversarial inputs and membership inference~\cite{papernot2017ModelExtraction}. While prior work has indicated the feasibility of model extraction attacks on various domains~\cite{papernot2017ModelExtraction, buseBoogeyman, szyller2021GANStealing, pal2019NLPModelStealing}, recent work introduced such attacks against GNNs as well~\cite{transductiveGNNStealAdversarial, transductiveGNNStealTaxonomy, shen2022GNNInductiveSteal}. 

Current approaches to address model extraction attacks rely on post-hoc \emph{ownership verification} in which the model owner requests a trusted verifier to decide whether a \emph{suspect model} was stolen from their target model. Ownership verification is done using either fingerprinting or watermarking. Watermarking has been shown to degrade model accuracy~\cite{taesun2019Perturbations, juuti2018PRADA, manish2018MEWarning} and can be evaded~\cite{lukas2021Conferrable, maini2021DatasetInference}. Hence, we identify fingerprinting as a potential scheme that can be used for model ownership verification. \emph{Model fingerprinting} uses inherent features of the model to distinguish between surrogates and independent models.~\cite{zhao2020FingerprintAdversarial, lukas2021Conferrable, cao2021FingerprintIPGUard, wang2021FingerprintDeepFool, peng2022FingerprintUAP, zheng2022FingerprintNonRepudiable}. On the other hand, \textit{dataset fingerprinting} uses the training data as a fingerprint, such that any model trained on the same data as the target model is classified as stolen~\cite{maini2021DatasetInference}. Such fingerprinting schemes have been proposed for non-graph DNNs, but there is currently no such work for GNNs. 

In this work, we present the \textbf{first fingerprinting scheme for GNNs}. We claim the following main contributions:
\begin{enumerate}[leftmargin=*]
\item identify GNN embeddings as a potential fingerprint and show that they are useful for verifying model ownership but not dataset ownership (Section~\ref{sec:motivation}). 
\item present \method, embedding-based fingerprinting for GNN model ownership verification (Section~\ref{sec:approach}).
\item extensively evaluate \method on six datasets and three architectures showing that \method is:
\begin{itemize}
    \item effective at distinguishing between surrogate and independent models with close to zero false positives or false negatives (Section~\ref{eval:effectiveness}), 
    \item robust against known fingerprint removal techniques (Section~\ref{eval:robustness}), and
    \item computationally efficient (Section~\ref{eval:efficiency}).
\end{itemize}
We open source our implementation: \url{https://github.com/ssg-research/GrOVe}
\end{enumerate}

\section{Background}\label{sec:background}

We describe some preliminaries for GNNs and notations used in this work (Section~\ref{back:gnn}), followed by an overview of model extraction attacks and defenses (Section~\ref{back:modelext}). 

\subsection{Graph Neural Networks}\label{back:gnn}

\begin{topics}
\topic{Graphs are a special type of data that has nodes and edges connecting these nodes.}
\topic{GNNs can be trained inductively and transductively. In the transductive setting, a GNN is trained on a semi-labeled graph to label the unlabelled nodes but cannot be used for unseen graphs. In the inductive setting, it is trained on fully labeled graphs and can be used on unseen graphs.}
\topic{GNNs learn an embedding for each node that is an aggregation of that node's features with its neighboring nodes.}
\end{topics}

Several real-world applications can be modeled as graphs that include nodes that represent different entities in the graph (e.g., authors in citation networks or users in social networks) and edges that represent the connections between nodes. Formally, a graph can be represented as $\mathcal{G} = (\mathcal{V}, \mathcal{E})$, where $\mathcal{V}$ is a set of nodes and $\mathcal{E}$ is a set of edges connecting these nodes. We represent a single node as $v \in \mathcal{V}$ and an edge between nodes $u$ and $v$ as $e_{uv} \in \mathcal{E}$. The entire graph structure, including all the nodes and corresponding edges, can be represented using a binary adjacency matrix $\mathcal{A}$ of size $|\mathcal{V}| \times |\mathcal{V}|$: $\mathcal{A}_{uv} = 1, \forall{(e_{uv})} \in \mathcal{E}$.

\noindent\textbf{\underline{ML on Graphs.}} Due to the large scale of these graph datasets, ML approaches for processing them have gained significant attention. Specifically, GNNs have shown tremendous performance in processing graph data for node and edge classification, clustering, and other tasks. Following prior work~\cite{shen2022GNNInductiveSteal, transductiveGNNStealAdversarial, transductiveGNNStealTaxonomy}, we consider node classification tasks in this work.

Each node has a feature vector $x \in \mathcal{X}$ and corresponding classification label $y \in \mathcal{Y}$ where $\mathcal{X}$ and $\mathcal{Y}$ are the set of features and labels across all nodes respectively.
We can denote the graph dataset as $\mathcal{D} = (\mathcal{A}, \mathcal{X}, \mathcal{Y})$ which is a tuple of the adjacency matrix, the set of node features, and the set of labels respectively. 

GNNs are trained to take the graph's adjacency matrix and the features as input and map it to the corresponding classification labels. Once trained, GNNs output a node embedding $h \in \mathcal{H}$. Embeddings are low-dimensional representations of each node and the graph structure and can be used for downstream tasks such as classification, recommendations, etc. We note that since \embds is dependent on the graph structure and the node features, two GNNs trained on different datasets should differ in their output of \embds~\cite{GraphSAGE, gat}. Formally, the mapping for a GNN is written as: $\mathcal{F}: \mathcal{A} \times \mathcal{X} \rightarrow \mathcal{H}$.

There are two training paradigms for GNNs:
\begin{itemize}[leftmargin=*]
    \item \textbf{Transductive} where the model trains on mapping some graph nodes to classification labels but uses the remaining graph nodes for prediction during testing. Here, the underlying graph structure $\mathcal{A}$, passed as input to the model, remains the same. This is useful for labeling a partially-labeled graph.
    \item \textbf{Inductive} where the model is trained on a training graph dataset but evaluated on an unseen and disjoint testing dataset.
\end{itemize}

Following prior work~\cite{shen2022GNNInductiveSteal}, we consider the more prevalent and practical setting of inductive training, which is applicable for ML as a service on graph data.

\noindent\textbf{\underline{GNN Computation.}} Training and evaluating a GNN involves aggregating information from neighboring nodes to compute the embeddings for a specific node. Formally, each layer of a GNN performs the following operation:
\begin{equation}
h_{v}^{l} = \mathrm{AGG}(h_{v}^{l-1}, \mathrm{MSG}(h_{v}^{l-1}, h_{u}^{l-1}: u \in \mathcal{N}(v)))
\end{equation}
$\mathcal{N}(v)$ denotes the nodes that share an edge with $v$. $h_{v}^{l}$ denotes the embedding of node $v$ at layer $l$. $h_{v}^{0}$ is initialized as the feature vector $x$ for node $v$. $\mathrm{MSG(\cdot)}$ gathers information from the neighbouring nodes of $v$, and $\mathrm{AGG(\cdot)}$ aggregates this information with $h_{v}^{l-1}$ to produce $h_{v}^{l}$. The primary difference between different GNN architectures is the implementation of these two functions. 

\noindent\textbf{GraphSAGE~\cite{GraphSAGE}} uses the mean aggregation operation:
\begin{equation}
h_{v}^{l} = \mathrm{CONCAT}(h_{v}^{l-1}, \mathrm{MEAN}(h_{u}^{l-1}: u \in \mathcal{N}(v)))
\end{equation}
where CONCAT is the concatenation operation, and MEAN is the mean operation.

\noindent\textbf{Graph Attention Networks (GAT)~\cite{gat}} includes masked self-attention layers in the GNN, which allows the model to assign a different weight to each neighbor of a node. This captures the variation in the contribution of different neighboring nodes. The aggregation function is:
\begin{equation}
h_{v}^{l} = \mathrm{CONCAT}_{k=1}^{K} \sigma (\sum_{u \in \mathcal{N}(v)} \alpha_{uv}^{k} \mathbf{W}^k h_{u}^{l-1})
\end{equation}
where CONCAT is the concatenation operation, $K$ is the total number of projection heads in the attention mechanism, $\alpha_{uv}^{k}$ is the attention coefficient in the $k^{th}$ projection head, $\mathbf{W}^k$ is the linear transformation weight matrix, and $\sigma(\cdot)$ is the activation function.

\noindent\textbf{Graph Isomorphism Network (GIN)~\cite{gin}} extends GraphSAGE and uses the aggregation function:
\begin{equation}
h_{v}^{l} = MLP^l ((1 + \epsilon^l) \cdot h_{v}^{l-1} + \sum_{u \in \mathcal{N}(v)} h_{u}^{l-1})    
\end{equation}
where MLP is a multi-layer perceptron and $\epsilon$ is a learnable parameter to adjust the weight of node $v$. This treats $h_{u}^{l-1}: u \in \mathcal{N}(v)$ as a \emph{multiset}, \textit{i.e.}, a set with possible repeating elements.

\subsection{Model Extraction Attacks and Defences}\label{back:modelext}

\begin{topics}
\topic{Model extraction attacks have been extensively studied in many domains and are considered a realistic threat.}
\topic{Previous work has described model extraction attacks against GNNs.}
\topic{Although various defenses have been discussed, they have been shown to be brittle [Nils]. Fingerprinting, a deterrence technique, appears to be the most promising in the image domain [Nils, DI]}
\topic{Defenses against model extraction in GNNs remain under-explored [REF]. In particular, there is no known fingerprinting scheme.}
\end{topics}

Model extraction attacks consider an adversary (\adv) who trains a local surrogate model (\surrogate) to mimic the functionality of a target model (\target)~\cite{tramer2016ModelExtraction}. These attacks have been extensively studied in many domains including images~\cite{okerondy2018ModelExtraction, papernot2017ModelExtraction, jagielskiHighFidHighAcc}, text~\cite{krishna2020NLPModelStealing, pal2019NLPModelStealing}, and graphs~\cite{shen2022GNNInductiveSteal, transductiveGNNStealTaxonomy, transductiveGNNStealAdversarial} and across different types of models including generative models~\cite{szyller2021GANStealing, hu2021GANStealing}, and large language models~\cite{wallace2020MTStealing}. This attack has been identified as a realistic threat that violates the confidentiality of the company's proprietary model and thereby their intellectual property~\cite{buseBoogeyman}. 

We denote the training dataset of \target (respectively \surrogate) as \targetdata (\surrogatedata). Additionally, we refer to models trained independently on a dataset \independentdata (in the absence of model extraction attack) are called independent models (\independent). 

\noindent\textbf{\underline{Non-Graph Model Extraction Attacks.}} \adv, given query access to \target, sends queries and obtains corresponding predictions. \adv then uses these input-prediction pairs to train \surrogate. Most attacks train \surrogate using specially crafted adversarial examples as inputs to \target ~\cite{tramer2016ModelExtraction, okerondy2018ModelExtraction, papernot2017ModelExtraction,jagielskiHighFidHighAcc}. This helps to ensure that the decision boundary of \surrogate is similar to \target. After a successful attack, \adv can use \surrogate to generate effective transferrable adversarial examples~\cite{papernot2017ModelExtraction} or perform membership inference attacks~\cite{shokri2017MIA}.

\noindent\textbf{\underline{Non-Graph Model Extraction Defenses.}} Preventing model extraction attacks without affecting model performance is difficult~\cite{buseBoogeyman, jagielsiCrypto, krishna2020NLPModelStealing}. However, ownership verification as a post-hoc approach helps identify whether a suspect model (\suspect) is stolen via model extraction. Normally, this involves a third-party verifier (\verifier) that verifies ownership. There are currently two main schemes for ownership verification:
\begin{itemize}[leftmargin=*]
    \item \textit{watermarking}~\cite{uchida2017Watermarking, wang2021CovertWatermarking, adi2018Backdooring, merrer2020Watermarking, Guo2018Watermarking, sablayrolles2020Radioactive, Li2021EmbeddedExternalFeatures, chen2019Watermarking, he2019Watermarking, szyller2021WatermarkingDawn} where some secret information is embedded into the model during training which is extracted later during verification.
    \item \textit{fingerprinting}~\cite{lukas2021Conferrable, maini2021DatasetInference, peng2022FingerprintUAP, cao2021FingerprintIPGUard, zhao2020FingerprintAdversarial, zheng2022FingerprintNonRepudiable} where inherent features are extracted from a model, without affecting the training process. 
\end{itemize}
Prior work has shown watermarking is brittle and can be easily evaded~\cite{lukas2021WatermarkingRobustness}. Hence, fingerprinting is currently the most promising technique for ownership verification. In this work, we focus on fingerprinting. 

Prior work has explored two fingerprinting schemes based on whether the fingerprint is for \emph{dataset ownership} or \emph{model ownership}.
The subtle difference between them is how a \emph{different} model trained from scratch \emph{on the same dataset} as \target is treated. Dataset-ownership-based fingerprinting classifies such a model as a surrogate. However, model-ownership-based fingerprinting classifies such a model as independent. Here, only models derived (e.g., transfer-learning, model extraction) from \target are classified as a surrogate. We describe the main prior works for fingerprinting below. 

For dataset ownership-based fingerprinting, \textbf{Maini et al.}~\cite{maini2021DatasetInference} propose \textit{dataset inference} on the following intuition: the distance of the training data points from the model's decision boundary is increased during training. Hence, the distance of a data point from the decision boundary can help infer its membership in \targetdata. \verifier queries \suspect with data points from \targetdata and unseen public data to compute their distances from the decision boundary. \suspect is classified as a surrogate if the distances corresponding to \targetdata are large, and the distances corresponding to unseen public data are small.

Most prior work on model ownership-based fingerprinting identifies adversarial examples that can transfer from \target to \surrogate but not to \independent~\cite{lukas2021Conferrable, zhao2020FingerprintAdversarial, cao2021FingerprintIPGUard, wang2021FingerprintDeepFool}. They vary in how to identify such adversarial examples. For instance, data points close to the decision boundary can be used to differentiate between \surrogate and \independent models~\cite{cao2021FingerprintIPGUard, wang2021FingerprintDeepFool}. Alternatively, untargeted adversarial examples can be used as well~\cite{zhao2020FingerprintAdversarial}. However, these works consider \surrogate derived using transfer-learning and fine-tuning but do not consider model extraction attacks~\cite{lukas2021Conferrable}.

Two prior model ownership-based fingerprinting works are evaluated explicitly with respect to model extraction attacks. We describe them below.

\textbf{Lukas et al.}~\cite{lukas2021Conferrable} use an ensemble of models to generate \emph{conferrable adversarial examples}, i.e., adversarial examples that transfer from \target to \surrogate, but are not misclassified by \independent. During verification, \verifier computes the error between the predictions of \target and \suspect on the conferrable examples. If the error rate exceeds a certain threshold, it is classified as independent, and surrogate otherwise.

\textbf{Peng et al.}~\cite{peng2022FingerprintUAP} use \emph{universal adversarial perturbations} (UAPs), small imperceptible perturbations which, when added to any image, result in misclassification. They compute a fingerprint by adding UAP to some data points and compute the change in output before and after UAP. They train an encoder to ensure that the fingerprints of \target and \surrogate are similar but distinct from \independent.

We explain in Section~\ref{prob:limitations} why these techniques (\cite{maini2021DatasetInference, lukas2021Conferrable, peng2022FingerprintUAP}) are not easily applicable to GNNs.


\noindent\textbf{\underline{Model Extraction Attacks on GNNs.}} There is limited literature on model extraction attacks in GNNs. One approach is to use adversarial examples to perform model extraction attacks, similar to the work in non-graph settings~\cite{transductiveGNNStealAdversarial}. However, the input perturbation for adversarial examples in graph setting is too high to be stealthy.
Wu et al.~\cite{transductiveGNNStealTaxonomy} presented seven attacks with different \adv background knowledge. However, these works are limited to transductive training, impractical in ML as a service setting~\cite{shen2022GNNInductiveSteal}.

We focus on the more practical case of inductive learning where the training and testing graph datasets are disjoint. Shen et al.~\cite{shen2022GNNInductiveSteal} presented the first work on model extraction against inductive GNNs. Here, \adv has access to a query dataset $\mathcal{D}_s = (\mathcal{A}_s, \mathcal{X}_s, \mathcal{Y}_s$) and the query response (a set of embeddings (\embdsettarget) corresponding to the node features ($\mathcal{X}_s$) it receives from \target. Using \embdsettarget and the ground-truth labels ($\mathcal{Y}_s$), \adv trains \surrogate to mimic the behavior of \target. They propose two attacks depending on \adv's background knowledge: in \emph{Type I} attack, \adv has access to $\mathcal{X}_s$, $\mathcal{Y}_s$, and the adjacency matrix ($\mathcal{A}_s$) for \surrogatedata, in \emph{Type II} attack, \adv only has access to $\mathcal{X}_s$, $\mathcal{Y}_s$, and uses an edge-estimation algorithm to compute $\mathcal{A}_s$. 

In their attack, the architecture for \surrogate consists of two components: 
\begin{itemize}[leftmargin=*]
\item a GNN that takes $\mathcal{X}_s$ and $\mathcal{A}_s$ as input and outputs \embdsetsurrogate. While training, this module minimizes the mean squared error (MSE) loss between \embdsetsurrogate and \embdsettarget:
\begin{gather}
    \mathcal{H}_s = \mathcal{G}nn (\mathcal{X}_s, \mathcal{A}_s) \\
    \mathcal{L}_R = \frac{1}{n_{Ds}} \|\mathcal{H}_s - \mathcal{H}_t \|_{2,1} \label{MSE}
\end{gather}
where $n_{Ds}$ is the number of nodes in \surrogatedata.
\item an MLP classifier (\classifier) which takes $\mathcal{H}_s$ as input and outputs a class. This is trained to minimize the prediction loss between $\mathcal{Y}_s$ and the predicted labels. 
\end{itemize}
In each epoch, $\mathcal{G}nn$ is first optimized using $\mathcal{L}_R$; then, the parameters are fixed while parameters corresponding to \classifier are optimized. Both modules are combined to form \surrogate.

The same training strategy is used to train \surrogate using Type I and Type II attacks. However, for Type II attacks, \adv first estimates $\mathcal{A}_s$. A graph structure is initialized by creating a $k$-nearest neighbors graph from $\mathcal{X}_s$. A search algorithm looks for hidden graph structure that augments the initial $k$-nearest neighbors graph~\cite{transductiveGNNStealTaxonomy}. The graph structure is obtained by minimizing a joint loss function that combines prediction loss for node classification and a graph regularization loss that controls the graph's smoothness, connectivity, and sparsity. Additional details about the model extraction attack can be found in~\cite{shen2022GNNInductiveSteal}.

\noindent\textbf{\underline{Model Extraction Defenses For GNNs.}} To the best of our knowledge, there are two prior works on watermarking in the context of GNNs~\cite{watermarkingGNNsBackdoor, watermarkGNNsRandomGraph}. Zhao et al.~\cite{watermarkGNNsRandomGraph} embed randomly generated subgraphs with random feature vectors as a key. These can be used later to extract the watermark. However, they only focus on node classification tasks. Xu et al.~\cite{watermarkingGNNsBackdoor} extend the prior work by including graph classification tasks. There are no known fingerprinting schemes for GNNs in the current literature.

\begin{table}[!htb]
\caption{Summary of notations used in this work.}
\label{table:notations}
\begin{center}
\begin{tabular}{ |c|c| } 
 \hline
 \textbf{Notation} & \textbf{Description}\\ 
 \hline
 $\mathcal{D} = (\mathcal{A}, \mathcal{X}, \mathcal{Y})$ & A graph dataset \\
 $\mathcal{A}$ & Adjacency matrix \\
 $\mathcal{X}$ & Set of node features $x$ \\
 $\mathcal{Y}$ & Set of labels $y$ \\
 $\mathcal{H}$ & Set of node embeddings $h$ \\
 $e_{uv}$ & An edge between nodes $u$ and $v$ \\
 $\mathcal{N}(v)$ & Neighbouring nodes of $v$ \\
 \target & Target model \\ 
 \surrogate & Surrogate model \\
 \independent & Independent Model \\
 \suspect & Suspect model to be verified \\
 \
 \targetdata~/ \surrogatedata~/ \independentdata & Target / Surrogate / Independent dataset \\
 \adv & Adversary \\
 \verifier & Verifier \\
 \accuser & Owner of \target \\
 \responder & Owner of \suspect \\
 \respadv & Malicious \responder \\
 \accadv & Malicious \accuser \\
 \committarget~/ \commitsuspect & Commitment of \target / \suspect \\
 \timetarget~/ \timesuspect & Timestamp of \target / \suspect \\ 
 \hline
\end{tabular}
\end{center}
\end{table}
\section{Problem Statement}\label{sec:problem}
Our goal is to design an effective fingerprinting scheme that allows \verifier to identify whether \suspect is a surrogate of \target or an independent model. 
We consider the state-of-the-art model extraction attacks in an inductive setting~\cite{shen2022GNNInductiveSteal} to evaluate our scheme. To this end, we outline a system model that defines the interactions between model owners and \verifier (Section~\ref{prob:sysmodel}), an adversary model describing \adv's capabilities and goals (Section~\ref{prob:advmodel}), requirements to design an ideal fingerprinting scheme (Section~\ref{prob:requirement}) and limitations of prior defenses against GNN model extraction (Section~\ref{prob:limitations}). Table~\ref{table:notations} summarizes the notations used in this work.

\subsection{System Model}\label{prob:sysmodel}

We consider a setting where a proprietary GNN model (\target) has been developed and deployed as a service. However, \target is susceptible to model extraction. 
An ownership verification system, intended to thwart model extraction, consists of three actors: an \accuser (the owner of \target), a trusted third party \verifier, and a \responder (the owner of a suspect model \suspect) who is accused of stealing \target by \accuser. The role of the \verifier is to verify whether \suspect was obtained through a model extraction attack on \target.  We refer to a malicious \responder as \respadv and a malicious \accuser as \accadv.

\noindent\textbf{\underline{System Design Goals.}} An ideal system must be robust against both \respadv and \accadv.

\begin{enumerate}[label=\textbf{Case \arabic*},leftmargin=*]
\item \respadv wants its model \suspect, extracted from \accuser's \target, to evade detection.
\item \accadv wants to maliciously claim that \responder's \suspect is extracted from \target. 
\end{enumerate}

To address both scenarios, similar to prior work~\cite{zheng2022FingerprintNonRepudiable, szyller2021WatermarkingDawn}, we assume that all model owners are required to securely timestamp their models in a registration step before deployment. This will address \accadv since \accuser cannot successfully make an ownership claim against \suspect unless \target was registered prior to \suspect. In Section~\ref{sec:discussions}, we explore possible incentives for model owners to register their models. In the rest of this paper, we will focus on robustness against a malicious \respadv.

\noindent\textbf{\underline{Model Registration.}} Model owners are required to:
\begin{enumerate}
\item generate a cryptographic commitment \commitment of their model  such that any subsequent modification to the model can be detected. One way to compute such a commitment is using a cryptographic hash function.
\item obtain a secure timestamp \timestamp on \commitment that \verifier can later verify, e.g.,  by adding \commitment to a blockchain, or utilizing a publicly verifiable timestamping service\footnote{E.g., https://www.surety.com/digital-copyright-protection}, or receiving a signature from \verifier binding \commitment to the current time.
\end{enumerate}

We use subscripts to associate timestamps and commitments to the respective models (e.g., \timetarget is the secure timestamp on the commitment \committarget of \target)

\noindent\textbf{\underline{Dispute Initiation.}} \accuser initiates a dispute by submitting \committarget and \timetarget to \verifier, and identifying a suspect \responder. \verifier then asks \responder to submit \commitsuspect and \timesuspect to begin the verification process. 

\noindent\textbf{\underline{Verification Process.}} \verifier does the following:
\begin{enumerate}
    \item verifies that \timetarget, \committarget, and \target are consistent; and \timesuspect, \commitsuspect, and \suspect are consistent.
    \item confirms that \timetarget < \timesuspect. If this condition is not met, the claim is rejected.
    \item \label{step-wellformed} checks that \target and \suspect are well-formed (see below).
    \item samples a verification dataset \inputverify from the same distribution as \targetdata.
    \item queries \target and \suspect with \inputverify and passes the outputs to a verification algorithm which decides whether \suspect is a surrogate of \target or trained independently.
\end{enumerate}
Step~\ref{step-wellformed} requires \verifier to check that a model does not have any non-standard layers. In Section~\ref{eval:robustness}, we explain why this check is necessary. 

The simplest way for \verifier to conduct the checks in the first three steps is for the model owners to send their models (\target and \suspect) to \verifier. This may not be feasible for confidentiality or privacy reasons. Thus, the checks can be done either via cryptographic techniques like oblivious inference~\cite{liu2017obliviousInference, juvekar2018obliviousInference} in conjunction with zero-knowledge proofs~\cite{kang2021ZKP, boneh2021ZKP} or by using hardware-based
trusted execution environments~\cite{tramer2018tee, deng2022tee}. The specific implementation of such protocols is out of the scope of this work. For ease of exposition, we limit our discussion to the case where the model owners in a dispute are willing to share their models with \verifier. Note that it is still necessary to check that the models sent to \verifier are indeed the models that were deployed. \verifier can do this via a fidelity check~\cite{jagielskiHighFidHighAcc}.

\subsection{Adversary Model}\label{prob:advmodel}
\begin{topics}
    \topic{We consider the black-box setting for \adv, so they only have access to the node embeddings. The verifier has block-box access to both the target model \target and the suspect model \suspect.}
    \topic{The goal of \adv is to build a surrogate model \surrogate that matches the behavior of \target}
    \topic{The goal of the verifier is to classify \suspect either as an independently trained model \independent or a surrogate model. \surrogate}
\end{topics}

\noindent\textbf{\underline{\respadv's goal}} is to train \surrogate such that its utility is comparable to \target. Additionally, \respadv wants high \emph{fidelity} for \surrogate, i.e., that its inferences match \target's. This is useful for mounting subsequent evasion or membership inference attacks against \target~\cite{papernot2017ModelExtraction}. Shen et al.'s~\cite{shen2022GNNInductiveSteal} attack satisfies both these requirements. \respadv may also take additional steps to evade detection.

\noindent\textbf{\underline{Attack Setting.}} Following \cite{shen2022GNNInductiveSteal}, we consider the \textit{black-box} setting where \respadv has no knowledge of \target's hyperparameters or architecture and can only observe \target's outputs for given inputs. 
As in prior work~\cite{shen2022GNNInductiveSteal}, we assume that the output contains node embeddings ($\mathcal{H}$), useful for downstream tasks such as classification, recommendation engines, visualizations, etc.
%
We assume that \respadv has access to a dataset \surrogatedata from the same distribution as \target's training data \targetdata. However, \surrogatedata and \targetdata are disjoint. We revisit the details of dataset splits in Section~\ref{sec:setup}.

\subsection{Requirements for Ownership Verification}\label{prob:requirement}
    \begin{topics}
    \topic{An ideal ownership verification system is non-invasive, effective, robust, and efficient.}
\end{topics}

We list desiderata for ideal GNN ownership verification schemes that \verifier can use to decide if \suspect is stolen from \target: 
\begin{enumerate}[label=\textbf{R\arabic*},leftmargin=*]
\item\label{req1} \textbf{Minimize Utility Loss} of \target.
\item\label{req2} \textbf{Effective} in differentiating between \surrogate and \independent with low false positive/negative rates. 
\item\label{req3} \textbf{Robust} in remaining effective against \respadv. 
\item\label{req4} \textbf{Efficient} by imposing a low computational overhead.
\end{enumerate}

\subsection{Limitations of Prior Work}\label{prob:limitations}
\begin{topics}
    \topic{Prior work on fingerprinting does not apply to GNNs since they assume each data point is independent. Each data point (node) in graphs depends on its neighboring nodes.}
    \topic{Prior work on ownership verification on GNNs uses watermarks, which violate the non-invasive property. They degrade the model performance on the primary task.}
\end{topics}

We now discuss how different prior works (Section~\ref{back:modelext}), applicable to non-graph and graph datasets, do not satisfy the above requirements.

\noindent\textbf{\underline{Non-Graph Datasets.}} Focusing only on approaches tested against model extraction attacks, we discuss how the three prior fingerprinting approaches for the image domain are not directly applicable to GNNs. We describe the limitations of prior non-graph fingerprinting schemes below.

Maini et al.~\cite{maini2021DatasetInference} compute the distance of a data point to the decision boundary by adding noise to the data points, which is not clear for inter-connected graph nodes~\cite{transductiveGNNStealAdversarial}. Moreover, prior works have indicated that dataset inference incurs false positives~\cite{Li2021EmbeddedExternalFeatures, szyller2022DIRobustness}. Lukas et al.~\cite{lukas2021Conferrable} and Peng et al.~\cite{peng2022FingerprintUAP} rely on adversarial examples as fingerprints. However, unlike images, generating adversarial examples is not trivial for graphs~\cite{zugner2018Adversarial, dai2018AdversarialAttack}.  

In summary, adapting non-graph fingerprinting approaches to graph datasets is not trivial, as data records in the image domain are independent. In contrast, nodes in graphs are related and satisfy homophily.  

\noindent\textbf{\underline{Graph Datasets.}} Watermarking schemes have been proposed previously for GNNs~\cite{watermarkGNNsRandomGraph, watermarkingGNNsBackdoor}. However, watermarks in non-graph datasets can be easily removed by model extraction attacks and are hence not robust~\ref{req3}~\cite{lukas2021WatermarkingRobustness}. Their effectiveness~\ref{req2} against state-of-the-art model extraction attacks is not clear. Finally, watermarks require modifying how the model is trained, violating the non-invasive requirement~\ref{req1}. Hence, in this work, we focus on using fingerprints for ownership resolution in GNNs instead of watermarking, which satisfies all the desirable requirements.

\section{Can GNN embeddings serve as fingerprints?}\label{sec:motivation}
\begin{topics}
    \topic{Node embeddings learned by a GNN are unique to an input graph}
    \topic{Influenced by training graph structure and node features}
    \topic{High-fidelity of model extraction would lead to similar embeddings}
    \topic{Ownership Verification can be done on either the model or the data.}
    \topic{We find that embeddings can be used for model ownership verification but not dataset ownership verification.}
\end{topics}

We now motivate our use of GNN embeddings as a potential fingerprinting scheme for GNNs against model extraction attacks. We then evaluate whether embeddings are helpful for model fingerprinting or dataset fingerprinting.

\noindent\textbf{\underline{Embeddings as Fingerprint for GNNs.}} Recall from Section~\ref{back:gnn} that two independently trained GNNs should differ in their output of \embds. Furthermore, the state-of-the-art model extraction attack against GNNs~\cite{shen2022GNNInductiveSteal} focuses on optimizing \emph{fidelity}, i.e., ensuring alignment in predictions between \surrogate and \target. Hence, \surrogate is likely to generate embeddings more similar to \target on the same input graph than \independent.
This intuition forms the basis of our fingerprinting scheme, \method, which uses node embeddings as a fingerprint to distinguish between \surrogate and \independent for ownership verification.
Note that \embds is inherent to GNN model computation. Hence, they do not affect the utility of the GNN, satisfying requirement~\ref{req1}.

\noindent\textbf{\underline{Model Ownership vs. Dataset Ownership.}} Having identified graph embeddings as a potential fingerprint, we want to verify whether they are helpful for model or dataset ownership. Recall from Section~\ref{back:modelext} that the difference in fingerprinting for model ownership and dataset ownership is in how models trained on the same training dataset are classified. If embeddings generated from two GNNs trained on the same dataset cannot be distinguished, then embeddings are useful as fingerprints for dataset ownership. On the other hand, if embeddings generated from \surrogate, regardless of the training data, can be distinguished, they are helpful as fingerprints for model ownership. We test this using t-SNE projections of the embeddings to visualize them for different model architectures and datasets~\cite{t-sne, peng2022FingerprintUAP}. We refer to embeddings generated from \target, \surrogate, and \independent as \embdsettarget, \embdsetsurrogate, and \embdsetindependent, respectively.

We describe our experimental setup to infer whether embeddings can be used for data or model ownership verification.

\noindent\textbf{Dataset Ownership.} We train two models using either different or the same architecture and training dataset. When the training datasets are different, we split the dataset into three sets: \modelonedata and \modeltwodata, and a verification dataset \inputverify. We use three different architectures for each model, leading to nine pair-wise combinations. We use six datasets, resulting in 54 model pairs with the same training datasets and 54 model pairs with differing training datasets. 

We pass \inputverify to both models to generate embeddings and visualize their t-SNE projections. The graphs for \coauthor are shown in Figure~\ref{fig:exp1_distances}, and the rest of the graphs are in Appendix~\ref{appendix:dataset_ownership}. We found that in every case, the t-SNE projections of the embeddings generated from \inputverify are distinguishable, even when the two models use the same dataset and architecture. This shows us that embeddings may not be useful as dataset fingerprints.

\begin{figure}[!htb]
    \centering
    \subfigure[Different Training Data]{\includegraphics[width=0.23\textwidth]{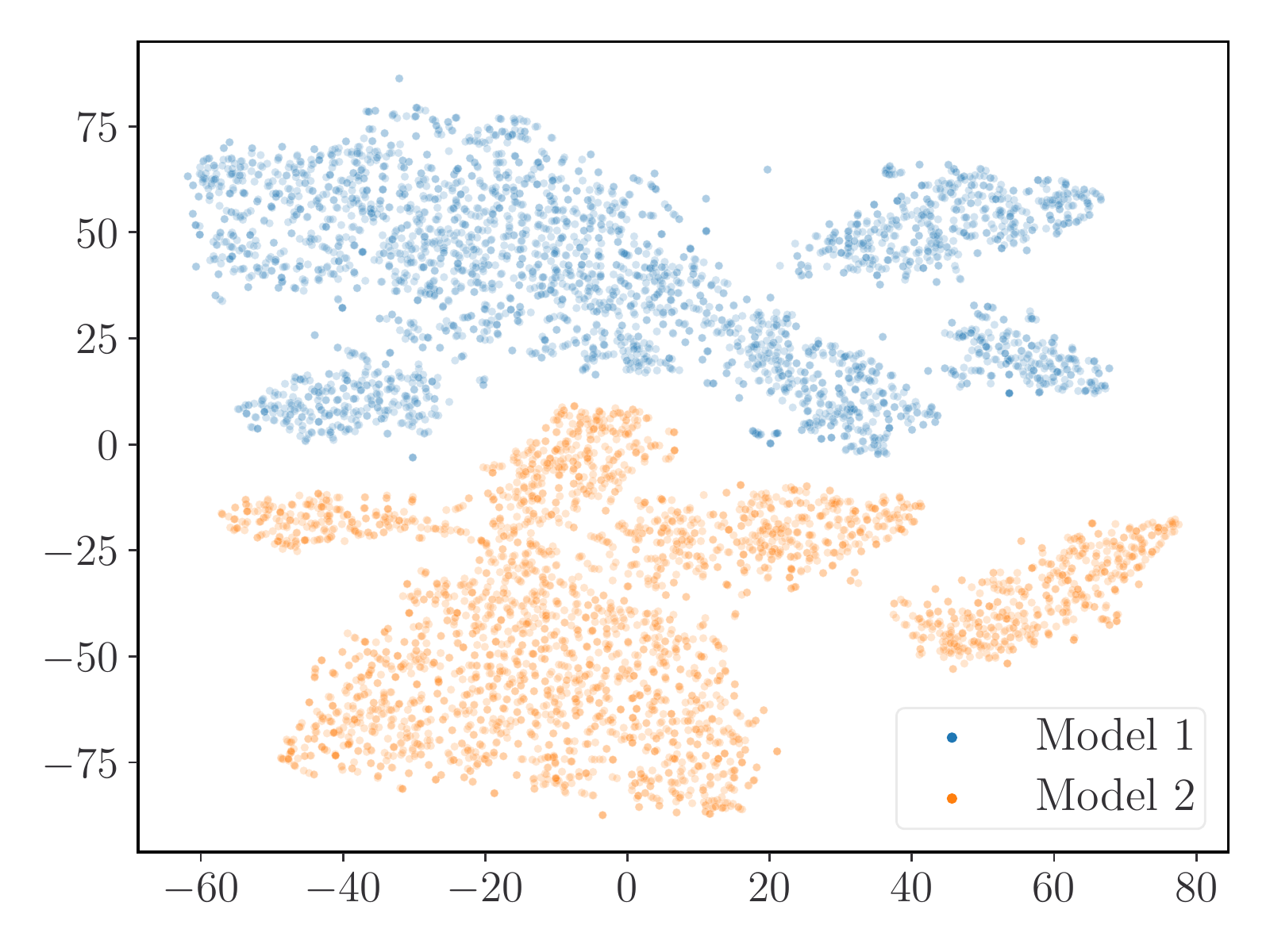}}
    \subfigure[Same Training Data]{\includegraphics[width=0.23\textwidth]{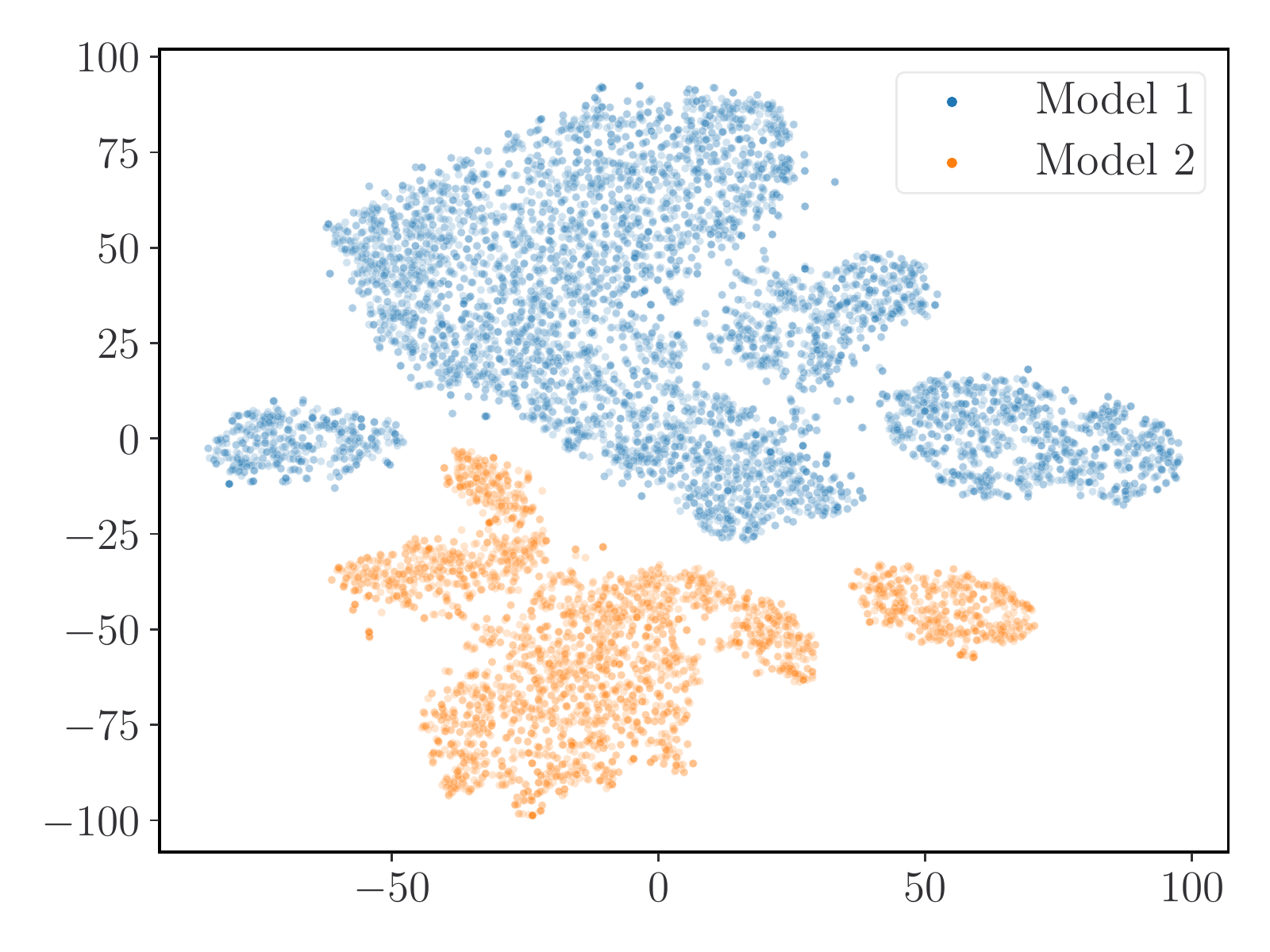}}
    \caption{t-SNE projections of the embeddings from two models trained on \coauthor using both the same and different architectures. The embeddings are distinguishable even when the architecture is the same. The graphs for the other datasets can be found in Appendix~\protect\ref{appendix:dataset_ownership}}
    \label{fig:exp1_distances}
\end{figure}

\noindent\textbf{Model Ownership.} To check for model ownership, we use three models: \target, \surrogate, \independent, which may or may not share the same architecture and training data as \target. We set up a similar experiment to the one before. We split the dataset into two training datasets: \modelonedata and \modeltwodata. \modelonedata is used to train both \target and \independent since this is the worst-case scenario for triggering false accusations using the fingerprinting scheme. \surrogate is derived from \target using \modeltwodata with Shen et al.'s~\cite{shen2022GNNInductiveSteal} state-of-the-art GNN model extraction attack described in Section~\ref{back:modelext}. Similar to the previous experiment, we build multiple combinations of the three models. We use three architectures for \target and \independent, and two for \surrogate, resulting in 108 model combinations across six datasets. 

We plot the embeddings of \inputverify from each of the three models. We show two of the graphs from \coauthor and \dblp in Figure~\ref{fig:exp2_distances} and the rest in Appendix~\ref{appendix_fig:model_ownership}. We found that regardless of the architecture or the training data, the t-SNE projections of \embdsettarget and \embdsetsurrogate are similar but distinct from \embdsetindependent. This motivates our choice to use embeddings as a fingerprint for model ownership. 

\begin{figure}[!htb]
    \centering
    \subfigure[Models trained on \coauthor dataset]{\includegraphics[width=0.35\textwidth]{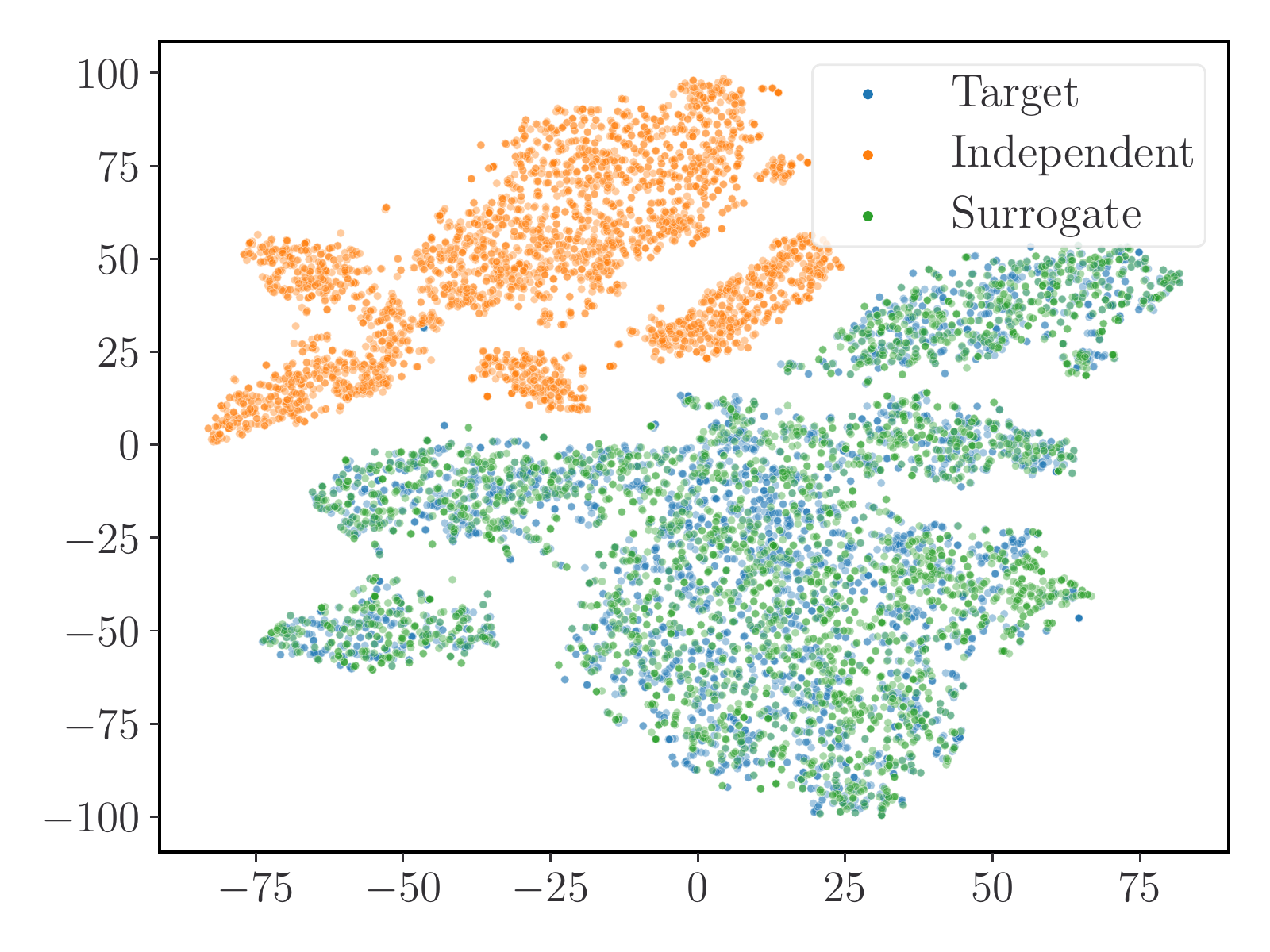}}\\
    \subfigure[Models trained on \dblp dataset]{\includegraphics[width=0.35\textwidth]{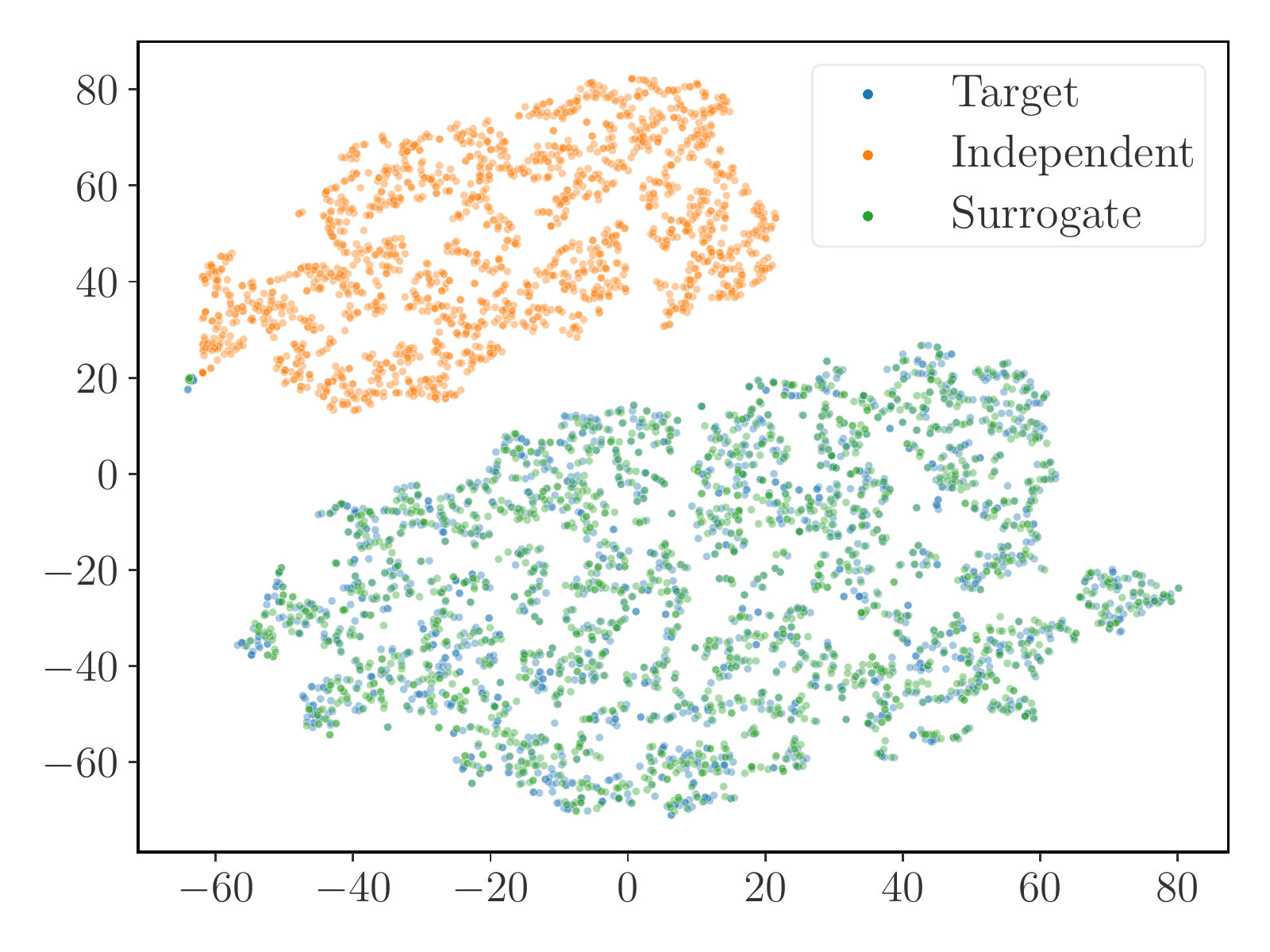}}\\
    \caption{t-SNE projections of the embeddings from \surrogate and \target overlap, while those from \independent are distinct. The models in this plot are all trained with the GAT architecture. The plots for the rest of the datasets are in Appendix~\protect\ref{appendix:model_ownership}}
    \label{fig:exp2_distances}
\end{figure}

\section{\method: Fingerprinting for GNN Ownership Verification}\label{sec:approach}
\definecolor{red1}{RGB}{221, 83, 84}
\definecolor{lightred1}{RGB}{231, 131, 131}
\definecolor{blue1}{RGB}{0, 108, 180}
\definecolor{darkblue1}{RGB}{0, 47, 108}
\definecolor{lightblue1}{RGB}{113, 198, 255}
\definecolor{green1}{RGB}{83, 205, 147}
\definecolor{orange1}{RGB}{227, 120, 40}
\definecolor{lightorange1}{RGB}{236, 162, 106}
\definecolor{yellow1}{RGB}{245, 203, 0}

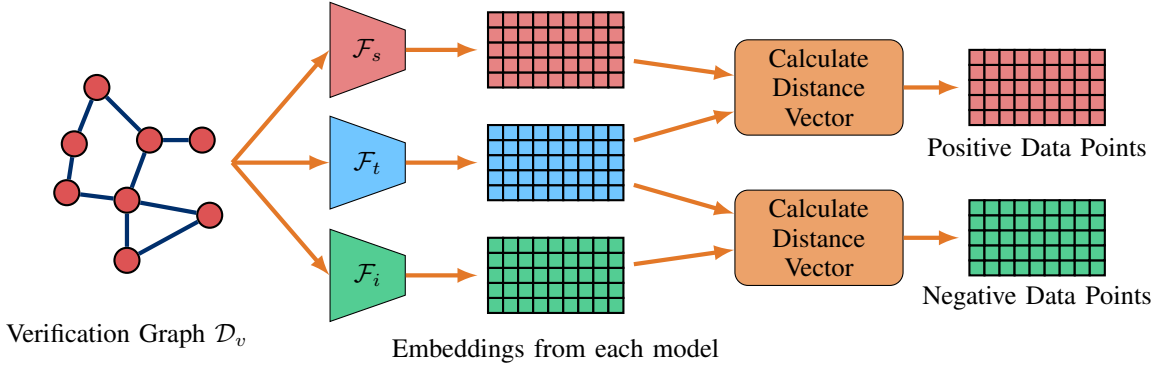
\begin{figure*}[h]
\begin{tikzpicture}

\begin{scope}[every node/.style={fill=red1, circle, thick, draw, shift={(-8,-0.5)}}]
    \node (1) at (-0.2,0) {};
    \node (2) at (-1,0.1) {};
    \node (3) at (0.1,0.8) {};
    \node (4) at (-0.6,1.5) {};
    \node (5) at (0.8,0.8) {};
    \node (6) at (0.9, -0.2) {};
    \node (7) at (-0.2,-0.8) {};
    \node (8) at (-0.9,0.75) {};
\end{scope}
\begin{scope}[ultra thick, darkblue1]
    \draw (2)--(1)--(3)--(4)--(8)--(2);
    \draw (3)--(5);
    \draw (1)--(6)--(7)--(1);
\end{scope}
\node [below of=7] {Verification Graph \inputverify};

\tikzset{
  model/.style={
    trapezium,
    draw=black,
    rotate=-90,
    trapezium stretches = true,
    minimum width=1cm,
    minimum height=1cm,
    fill=#1
  }
}
\node[model=lightblue1] (target) at (-5,0) {\rotatebox{90}\target};
\node[model=lightred1, left of=target, xshift=-0.5cm] (surrogate) at (-5,0) {\rotatebox{90}\surrogate};
\node[model=green1, right of=target, xshift=0.5cm] (independent) at (-5,0) {\rotatebox{90}\independent};

\draw[-latex, ultra thick, draw=orange1] (-6.8,0)--(surrogate.south);
\draw[-latex, ultra thick, draw=orange1] (-6.8,0)--(target.south);
\draw[-latex, ultra thick, draw=orange1] (-6.8,0)--(independent.south);

\node [right of=surrogate, node distance=2.5cm] (surrogateMatrix){
    \begin{tikzpicture}
        \foreach \x in {0,0.2,...,1.6} {
            \foreach \y in {0,0.2,...,0.8} {
                \draw[thick, black, fill=lightred1] (\x,\y) rectangle ++(0.2,0.2);
            }
        }    
    \end{tikzpicture}
};
\node [right of=target, node distance=2.5cm] (targetMatrix){
    \begin{tikzpicture}
        \foreach \x in {0,0.2,...,1.6} {
            \foreach \y in {0,0.2,...,0.8} {
                \draw[thick, black, fill=lightblue1] (\x,\y) rectangle ++(0.2,0.2);
            }
        }    
    \end{tikzpicture}
};
\node [right of=independent, node distance=2.5cm] (independentMatrix){
    \begin{tikzpicture}
        \foreach \x in {0,0.2,...,1.6} {
            \foreach \y in {0,0.2,...,0.8} {
                \draw[thick, black, fill=green1] (\x,\y) rectangle ++(0.2,0.2);
            }
        }    
    \end{tikzpicture}
};

\node[below of=independentMatrix]{Embeddings from each model};

\draw[-latex, ultra thick, draw=orange1] (surrogate.north)--(surrogateMatrix);
\draw[-latex, ultra thick, draw=orange1] (target.north)--(targetMatrix);
\draw[-latex, ultra thick, draw=orange1] (independent.north)--(independentMatrix);

\node[draw, rounded corners=5pt, fill=lightorange1, minimum width=2cm, minimum height=1cm, text width=2cm, align=center] (positiveDistance) at (1,1) {Calculate Distance Vector};
\node[draw, rounded corners=5pt, fill=lightorange1, minimum width=2cm, minimum height=1cm, text width=2cm, align=center] (negativeDistance) at (1,-1) {Calculate Distance Vector};

\draw[-latex, ultra thick, draw=orange1] (surrogateMatrix)--(positiveDistance);
\draw[-latex, ultra thick, draw=orange1] (targetMatrix)--(positiveDistance);
\draw[-latex, ultra thick, draw=orange1] (targetMatrix)--(negativeDistance);
\draw[-latex, ultra thick, draw=orange1] (independentMatrix)--(negativeDistance);

\node [right of=positiveDistance, xshift=0.4cm, node distance=2.5cm] (positiveSamples){
    \begin{tikzpicture}
        \foreach \x in {0,0.2,...,1.6} {
            \foreach \y in {0,0.2,...,0.8} {
                \draw[thick, black, fill=lightred1] (\x,\y) rectangle ++(0.2,0.2);
            }
        }    
    \end{tikzpicture}
};
\node[below of=positiveSamples, yshift=0.2cm] (positiveText) {Positive Data Points};

\node [right of=negativeDistance, xshift=0.4cm, node distance=2.5cm] (negativeSamples){
    \begin{tikzpicture}
        \foreach \x in {0,0.2,...,1.6} {
            \foreach \y in {0,0.2,...,0.8} {
                \draw[thick, black, fill=green1] (\x,\y) rectangle ++(0.2,0.2);
            }
        }    
    \end{tikzpicture}
};
\node[below of=negativeSamples, yshift=0.2cm] (negativeText) {Negative Data Points};

\draw[-latex, ultra thick, draw=orange1] (positiveDistance)--(positiveSamples);
\draw[-latex, ultra thick, draw=orange1] (negativeDistance)--(negativeSamples);

\end{tikzpicture}
\caption{To generate training data for \similarity, an input ``verification'' graph is passed to \target, \surrogate and \independent. We then compute the distance vectors between the embeddings obtained from each of the models, namely, (\target, \surrogate) and (\target, \independent) which act as positive and negative data points respectively.}
\label{fig:training}
\end{figure*}
\begin{topics}
    \topic{There are two phases in the verification scheme.}
    \topic{In the first phase, we train a similarity model (\similarity) that, given a pair of embeddings, outputs whether the pair of embeddings are close or far.}
    \topic{In the second phase, we query \target and \suspect with a verification graph \inputverify to get multiple pairs of embeddings. These embeddings are input to the similarity model. If more than 50\% of the pairs are classifier as close, we classify \suspect as \surrogate, and \independent otherwise.}
\end{topics}

We now describe the design of \method, which uses embeddings as fingerprints.

Recall that \verifier aims to identify whether \suspect was obtained via a model extraction attack on \target (Section~\ref{prob:sysmodel}). As shown in Section~\ref{sec:motivation}, we know that the embeddings generated by \target and \surrogate are similar. Our verification scheme relies on this observation and identifies whether the distances between \embdsetsuspect (generated by \suspect) and \embdsettarget are \emph{close enough} to suggest a model extraction attack. 

The simplest way to identify this is to calculate a distance metric between \embdsetsuspect and \embdsettarget. If the aggregated distances are smaller than a tuned threshold, we can classify \suspect and \surrogate, and \independent otherwise. However, our experiments found that the distances between (\target, \independent) pairs and (\target, \surrogate) pairs overlapped, leading to high error rates. This phenomenon is visualized in the distance plots in Appendix~\ref{app:distances}. 

Thus, similar to the intuition used for Siamese Networks~\cite{koch2015siamese}, we use an ML classifier to classify a pair of embeddings as \emph{similar} or \emph{not similar}. An ML-based approach compresses the distance between similar data points and stretches the distance between dissimilar data points. The ML classifier is a multi-layer perceptron, denoted as \similarity. Our verification scheme is divided into two phases: Phase 1 involves training \similarity; Phase 2 includes querying \similarity with pairs of embeddings generated by \suspect and \target on \inputverify to classify \suspect as either a surrogate or independent model. 

\noindent\textbf{\underline{\similaritybold Training.}} Figure~\ref{fig:training} shows the process of generating training data for \similarity. For each \target, we train multiple sets of \surrogate and \independent using different architectures. \verifier queries \target, \surrogate and \independent with \inputverify to generate embeddings. Each node in \inputverify will thus have three corresponding embeddings from each GNN, denoted as \embdtarget, \embdsurrogate, and \embdindependent, respectively. \verifier generates a distance vector between \embdtarget and \embdsurrogate, representing a positive (similar) data point. In contrast, the distance vector between \embdtarget and \embdindependent is represented as a negative (not similar) data point. The distance vector is the element-wise squared distance between the two embeddings. This allows \verifier to generate many data points from one pair of models equal to the size of \inputverify. \verifier then uses this data to optimize \similarity to classify whether a pair of embeddings is similar. 

\noindent\textbf{\underline{Verification.}} Once \similarity has been trained \verifier can use it to make inferences about \suspect. \verifier queries both \suspect and \target with \inputverify to generate embeddings. The distance vector is calculated between corresponding embeddings and passed to \similarity, which outputs whether the pair is similar. If more than $50\%$  of the pairs of embeddings are classified as similar, \verifier classifies \suspect as a surrogate, and independent otherwise. In our experiments we found that the precise threshold does not matter. In general, for (\target, \surrogate) pairs, $\approx90\%$ of the embeddings are similar, while for (\target, \independent) pairs, only $\approx10\%$ embeddings are classified as similar.

\section{Experimental Setup}\label{sec:setup}

We evaluate \method using six datasets, three model architectures, and two model extraction attacks. We describe the datasets and their splits (Section~\ref{setup:datasets}), model architectures (Section~\ref{setup:architectures}), metrics for evaluation (Section~\ref{setup:metrics}), and the model extraction attacks (Section~\ref{setup:attack}).

\subsection{Datasets}\label{setup:datasets}
\begin{topics}
    \topic{We use six datasets. DBLP~\cite{dblp}, Citeseer~\cite{citeseer}, Pubmed~\cite{pubmed}, Coauthor~\cite{coauthor}, ACM~\cite{acm}, and Amazon~\cite{amazon}.}
    \topic{35\% of the data is used to train \target, 35\% non-overlapping data is used to train \surrogate. Each model is evaluated on 10\% non-overlapping data. The remaining 10\% is used as \inputverify.}
    \topic{\similarity is trained by gathering pairs of embeddings from \target and multiple combinations of \surrogate and \independent. Some of the model combinations are used for training, others for testing.}
\end{topics}

\begin{short}

Following prior work~\cite{shen2022GNNInductiveSteal}, we consider six benchmark graph datasets representing different types of graph networks. We use three citation networks; DBLP~\cite{dblp}, Citeseer~\cite{citeseer}, and Pubmed~\cite{pubmed}, two co-authorship networks; Coauthor Physics, abbreviated as Coauthor~\cite{coauthor}, and ACM~\cite{acm}; and the Amazon Co-Purchase network, abbreviated as Amazon~\cite{amazon}.

\end{short}

Following prior work~\cite{shen2022GNNInductiveSteal}, we consider six benchmark graph datasets representing different types of graph networks. We describe the details of each of these datasets below.

\noindent\textbf{\dblp}~\cite{dblp} is a citation network where the 17,716 nodes represent publications and 105,734 edges indicate citations between different publications. Each node has 1,639 features based on the keywords in the paper. This is a node classification problem with four classes indicating the publication category.

\noindent\textbf{\citseer}~\cite{citeseer} is a citation network where the 4,120 nodes represent publications and 5,358 edges indicate citations between different publications. Each node has 602 features indicating the absence/presence of the corresponding word from the dictionary. This is a node classification task where the publications are categorized into six classes.

\noindent\textbf{\pubmed}~\cite{pubmed} is a citation network where the 19,717 nodes represent publications and 88,648 edges indicate citations between different publications. Each node has 500 features described by a TF/IDF weighted word vector from a dictionary which consists of 500 unique words. This is a node classification task where the publications are categorized into three classes.

\noindent\textbf{\coauthor}~\cite{coauthor} is a co-authorship network where the 34,493 nodes represent different authors, which are connected with 495,924 edges if they coauthored a paper. Here, the 8,415 node features represent paper keywords for each author’s papers. This node classification task is to predict the most active field of study out of five possibilities for each author.

\noindent\textbf{\acm}~\cite{acm} is a heterogeneous graph that contains 3025 papers published in KDD, SIGMOD, SIGCOMM, MobiCOMM, and VLDB. Papers that the same author publishes have an edge between them, resulting in 26,256 edges. Each paper is divided into three classes (Database, Wireless Communication, Data Mining), and the 1,870 features for each node are the bag-of-words representation of their keywords.

\noindent\textbf{\amazon}~\cite{amazon} is an abbreviation for Amazon Co-purchase Network for Photos. The 7,650 nodes represent items, and the 143,663 edges indicate whether the two items are bought together. Each of the 745 nodes features are bag-of-words encoded product reviews. Each of the items is classified into one of eight product categories.

\noindent\underline{\textbf{Dataset Splits for Model Extraction.}} We split the dataset into two disjoint sets, each 40\% of the dataset. One chunk is used to train \target, and the other chunk is used to train \surrogate. While the original model extraction attacks use overlapping and non-overlapping training data for \surrogate, we choose a non-overlapping dataset since that is the most challenging case for \verifier, as \surrogate is distinct from \target. For the same reason, \target and \independent are trained on the same set. We evaluate \target and \surrogate on a test set that is 10\% of the dataset. Finally, \inputverify is the remaining 10\%. The dataset sizes are shown in Table~\ref{tab:sizes}

\begin{table}[h]
\caption{Data splits for training and evaluating different models.}
\centering
\begin{tabular}{ |c|c|c|c|c| } 
 \hline
\textbf{Dataset} & \textbf{\target Train} & \textbf{\surrogate Train} & \textbf{Test} & \textbf{\inputverify} \\
\hline
\textbf{\coauthor} & 13797 & 13797 & 3449 & 3449 \\
\textbf{\pubmed} & 7887 & 7887 & 1972 & 1972 \\
\textbf{\dblp} & 7086 & 7086 & 1772 & 1772 \\
\textbf{\amazon} & 3060 & 3060 & 765 & 765 \\
\textbf{\citseer} & 1648 & 1648 & 412 & 412 \\
\textbf{\acm} & 1210 & 1210 & 303 & 303 \\
\hline
\end{tabular}
\label{tab:sizes}
\end{table}

\subsection{Model Architectures}\label{setup:architectures}
\begin{topics}
    \topic{We use the GIN, GAT, and SAGE architectures for our models.}
\end{topics}
Following prior work~\cite{shen2022GNNInductiveSteal}, we use the GAT, GIN, and GraphSAGE architectures (Section~\ref{back:gnn}). For all architectures, the hidden layer size is 256, and the final hidden layer is connected to a dense layer for classification. The embeddings are extracted from the last hidden layer. We use the cross-entropy loss as the node classification loss, the ReLU activation function, and the Adam optimizer with an initial learning rate of 0.001. All models are trained for 200 epochs with early stopping based on validation accuracy. 

\noindent\textbf{GIN.} We use a three-layer GIN model. The neighborhood sample size is fixed at ten samples at each layer. 

\noindent\textbf{GAT.} We use a three-layer GAT model with a fixed neighborhood sample size of 10 at each layer. The first and second layers have four attention heads each. 

\noindent\textbf{GraphSAGE.} We use a
two-layer GraphSAGE model. The neighborhood sample sizes are set to 25 and 10, respectively. Following prior work~\cite{GraphSAGE}, the MEAN aggregation function is used at each layer, and the dropout is set to 0.5 to prevent overfitting.

\noindent\textbf{Similarity Model.} We use the Scikit-Learn implementation of an MLPClassifier to train \similarity. We train three versions of \target for each dataset using the three architectures. Each \target has its own \similarity. We train three versions of \independent and two versions of \surrogate for each \target which we use to generate the positive and negative data points as explained in Section~\ref{sec:approach}. We use a two-layer MLP and find the best hyperparameters using a grid search. The hidden layer sizes in the search are either 64 or 128, and the activation function is either Tanh or ReLU. We select the best model based on 10-fold cross-validation. We only train \similarity on Type I attacks unless otherwise stated.

To evaluate \similarity, we train nine different versions of \independent and \surrogate, i.e., 45 test models with different random initialization. These additional models are used to ensure the statistical significance of the results. We ran each experiment five times and reported the average with a 95\% confidence interval.

\subsection{Metrics}\label{setup:metrics}
\begin{topics}
    \topic{We use accuracy and fidelity to evaluate \surrogate. We use False Positive Rate (FPR) and False Negative Rate (FNR) to evaluate \method}
\end{topics}
Following prior work~\cite{shen2022GNNInductiveSteal,jagielskiHighFidHighAcc}, we use two metrics for evaluating the effectiveness of model extraction attacks: accuracy and fidelity.

\noindent\textbf{Accuracy} measures the number of predictions \surrogate classifies correctly, compared to the ground-truth. 

\noindent\textbf{Fidelity} measures the label agreement between \target and \surrogate. 

To evaluate the effectiveness of \method, we use two additional metrics:

\noindent\textbf{False Positive Rate (FPR)} is the ratio of false positives to the sum of false positives and true negatives. This indicates the fraction of independent models incorrectly classified as surrogate. 

\noindent\textbf{False Negative Rate (FNR)} is the ratio of false negatives to the sum of false negatives and true positives. This indicates the fraction of surrogate models incorrectly classified as independent.

\subsection{Model Extraction Attack}\label{setup:attack}
\begin{topics}
    \topic{We use the model stealing attack from prior work \cite{shen2022GNNInductiveSteal}}
    \topic{The original paper has two types of attacks, \emph{Type I} and \emph{Type II}. Type I attacks assume the adversary has access to a dataset \surrogatedata with similar distribution to \targetdata. Type II attacks assume the adversary only has access to nodes, not the edges, and thus needs to use an edge-estimation algorithm to run the model extraction attack.}
\end{topics}

We use Shen et al.'s~\cite{shen2022GNNInductiveSteal} model extraction attack against inductive GNNs from their source code: \url{https://github.com/xinleihe/GNNStealing}. The details of the attack have been explained in Section~\ref{back:modelext}. As mentioned in Section~\ref{setup:datasets}, we train \surrogate on \surrogatedata that is disjoint from \targetdata. All the hyperparameters are set to the default values presented in the original paper. The performance of these models is summarized in Table~\ref{tab:acc_fid}, and our results are similar to those reported in the original attack paper. 

\begin{table*}[!htb]
\caption{Average accuracy and fidelity values (with 95\% confidence intervals) of \target, \independent, and \surrogate used in the evaluation. Values are averaged across multiple architectures (Section~\protect\ref{setup:architectures}).}
\begin{center}
\begin{tabular}{|c|c|c|c|c|c|c|}
\hline
\textbf{Dataset} & \target \textbf{Accuracy} & \independent \textbf{Accuracy} & \textbf{Type I} \surrogate \textbf{Accuracy} & \textbf{Type I} \surrogate \textbf{Fidelity} & \textbf{Type II} \surrogate \textbf{Accuracy} & \textbf{Type II} \surrogate \textbf{Fidelity} \\
\hline
\textbf{\acm} & $0.906 \pm 0.025$ & $0.919 \pm 0.021$ & $0.888 \pm 0.019$ & $0.931 \pm 0.019$ & $0.896 \pm 0.010$ & $0.954 \pm 0.020$ \\
\textbf{\amazon} & $0.879 \pm 0.064$ & $0.876 \pm 0.050$ & $0.861 \pm 0.022$ & $0.870 \pm 0.051$ & $0.842 \pm 0.007$ & $0.848 \pm 0.009$ \\
\textbf{\citseer} & $0.804 \pm 0.047$ & $0.809 \pm 0.028$ & $0.757 \pm 0.014$ & $0.907 \pm 0.041$ & $0.796 \pm 0.000$ & $0.902 \pm 0.012$ \\
\textbf{\coauthor} & $0.926 \pm 0.005$ & $0.928 \pm 0.011$ & $0.919 \pm 0.019$ & $0.949 \pm 0.034$ & $0.919 \pm 0.004$ & $0.948 \pm 0.003$ \\
\textbf{\dblp} & $0.696 \pm 0.028$ & $0.693 \pm 0.030$ & $0.674 \pm 0.009$ & $0.833 \pm 0.018$ & $0.680 \pm 0.008$ & $0.851 \pm 0.017$ \\
\textbf{\pubmed} & $0.846 \pm 0.022$ & $0.846 \pm 0.021$ & $0.829 \pm 0.007$ & $0.923 \pm 0.016$ & $0.832 \pm 0.005$ & $0.937 \pm 0.014$ \\
\hline
\end{tabular}
\label{tab:acc_fid}
\end{center}
\end{table*}

\section{Evaluation of \method}\label{sec:evaluation}

We show that \method satisfies the requirements outlined in Section~\ref{prob:requirement}. To this end, we evaluate \method's effectiveness (requirement \ref{req2}) in differentiating between \surrogate, derived from the two types of model extraction attacks, and \independent (Section~\ref{eval:effectiveness}). We evaluate \method's robustness (requirement \ref{req3}) to \respadv's attempts at evading verification (Section~\ref{eval:robustness}). Finally, we evaluate \method's efficiency (requirement \ref{req4}) (Section~\ref{eval:efficiency}).

\subsection{Effectiveness}\label{eval:effectiveness}
\begin{topics}
    \topic{\method is able to distinguish between \surrogate and \independent effectively.}
\end{topics}

\begin{table}[!htb]
\caption{Performance of \method against Type 1 and Type 2 attacks. Average FPR and FNR values are reported across 5 experiments with 95\% confidence intervals.}
\centering
\begin{tabular}{|c|c|c|c|}
\hline
\textbf{Dataset} & \textbf{FPR} & \textbf{Type 1 FNR} & \textbf{Type 2 FNR} \\
\hline
\textbf{\acm} & $0.022 \pm 0.022$ & $0.000 \pm 0.000$ & $0.000 \pm 0.000$ \\
\textbf{\amazon} & $0.034 \pm 0.029$ & $0.000 \pm 0.000$ & $0.000 \pm 0.000$ \\
\textbf{\citseer} & $0.000 \pm 0.000$ & $0.000 \pm 0.000$ & $0.000 \pm 0.000$ \\
\textbf{\coauthor} & $0.000 \pm 0.000$ & $0.000 \pm 0.000$ & $0.000 \pm 0.000$ \\
\textbf{\dblp} & $0.000 \pm 0.000$ & $0.000 \pm 0.000$ & $0.000 \pm 0.000$ \\
\textbf{\pubmed} & $0.002 \pm 0.002$ & $0.000 \pm 0.000$ & $0.000 \pm 0.000$ \\
\hline
\end{tabular}
\label{tab:simple_fpr_fnr}
\end{table}

We first show that \method is effective at distinguishing between \independent and \surrogate (requirement \ref{req2}). To this end, we train three \independent models and two \surrogate models using the Type 1 attack to train \similarity. During the testing phase of \similarity, we train additional \independent models to compute the FPR and \surrogate models from both Type 1 and Type 2 attacks to compute the FNR.

Ideally, we expect \method to have low FPR and FNR while differentiating between \independent and \surrogate using \similarity. As seen in Table~\ref{tab:simple_fpr_fnr}, we find that \method indeed has zero false negatives against both model extraction attacks. We observe close to zero false positives across four datasets (\dblp, \pubmed, \citseer, and \coauthor). For the \amazon and \acm datasets, we note a low false positive rate of 3.4\% and 2.2\%, respectively.

These results show that \method is effective at distinguishing between \surrogate and \independent across different datasets and different architectures, satisfying \ref{req2}.

\subsection{Adversarial Robustness of \method}\label{eval:robustness}
\begin{topics}
    \topic{We test two methods to evade detection; \emph{double extraction} and \emph{model pruning}. The primary goal in both techniques is to change the distribution of the embeddings in an attempt to fool the verification model.}
    \topic{In double extraction, the adversary creates a surrogate model of the surrogate model. In model pruning, the adversary sets a ratio of the weights to zero so they do not contribute to the output.}
    \topic{We achieve consistently good results against double extraction.}
    \topic{We found that our original scheme did not perform well against model pruning. By including the embeddings from pruned models as additional training data for \similarity, our results improved.}
    \topic{\method is flexible enough to include additional attacks as part of the training data, including ones we have not evaluated.}
\end{topics}

We now discuss the robustness of \method against attempts to evade detection by \respadv (requirement \ref{req3}). \respadv can evade detection by differentiating \surrogate from \target via (1) simple evasion or (2) model retraining. 

\noindent\textbf{\underline{Simple evasion}} techniques can be used by \respadv to evade detection without retraining \surrogate: (a) \emph{model replacement} and (b) \emph{post-processing}. These are described below.

\noindent\textbf{Model replacement} 
occurs when during the verification process,
\verifier intends to query the model that \responder has deployed (\surrogate), but \respadv replaces it by an independent model \independent to deceive \verifier. 
Our system model (Section~\ref{prob:sysmodel}) pre-empts this by requiring \responder to register \surrogate before deployment. During verification, \verifier conducts a fidelity check between the outputs of the registered and deployed models (i.e., both outputs should match perfectly) to confirm that they are the same. As both models should be identical, the fidelity score between embeddings of the same input should be perfect.

\noindent\textbf{Post-processing} occurs when \respadv applies a (linear) transformation on \embdsetsurrogate to change its distribution while maintaining utility. If the distances between the embeddings of any two nodes of an input graph remain relatively the same after the transformation, it will not affect the result of any downstream task. Our system model pre-empts this by requiring \respadv to send \surrogate to \verifier, who verifies whether the outputs are generated directly from \surrogate without being post-processed. Furthermore, \verifier can inspect \surrogate to ensure there are no non-standard layers that arbitrarily transform the output (Step~\ref{step-wellformed} of the Verification Process in Section~\ref{prob:sysmodel}). 

\noindent\textbf{\underline{Model retraining.}} We identify four possible techniques from prior work to evade detection via model retraining~\cite{lukas2021Conferrable, peng2022FingerprintUAP}: (a) \emph{fine-tuning}, (b) \emph{double extraction}, (c) \emph{pruning} and (d) \emph{distribution-shift}. 

\noindent\textbf{Fine-tuning} retrains a previously trained model on a new dataset to improve model performance or change the classification task by replacing the model's final layer. While this is popular in prior work~\cite{lukas2021Conferrable, peng2022FingerprintUAP, maini2021DatasetInference}, we argue that it cannot be used by \respadv to evade detection of extracted GNN models. Recall from Section~\ref{setup:attack} that \surrogate consists of two independent components: the first outputs embeddings, and the second outputs class labels. Recalling \eqref{MSE}, the embeddings are updated using an MSE loss with embeddings from \target. Traditional fine-tuning based on different class labels will only update the classifier, not affecting the embeddings. 

Thus, we only test \method against end-to-end fine-tuning, where \respadv updates both the described components while fine-tuning on a separate dataset from the same distribution. The average performance of the models is reported in Table~\ref{tab:fine_tuning_performance}. We found that \surrogate accuracy improves slightly after end-to-end fine-tuning while the fidelity slightly decreases. Despite this change, \method still achieves zero false negatives across all datasets, showing \method is effective in mitigating end-to-end fine-tuning attacks across different datasets.

\begin{table*}[!htb]
\centering
\caption{Performance of end-to-end fine-tuning. In many cases \surrogate accuracy is higher, but the fidelity is lower. For \coauthor the accuracy of \surrogate surpassed \target after fine-tuning.}
\begin{tabular}{|c|c|c|c|c|}
\hline
\textbf{Dataset} & \textbf{Type 1} \surrogate \textbf{Accuracy} & \textbf{Type 1 }\surrogate \textbf{Fidelity} & \textbf{Type 2} \surrogate \textbf{Accuracy} & \textbf{Type 2} \surrogate \textbf{Fidelity} \\
\hline
\textbf{\acm} & $0.899 \pm 0.030$ & $0.927 \pm 0.041$ & $0.902 \pm 0.013$ & $0.939 \pm 0.021$ \\
\textbf{\amazon} & $0.875 \pm 0.018$ & $0.874 \pm 0.029$ & $0.866 \pm 0.021$ & $0.863 \pm 0.032$ \\
\textbf{\citseer} & $0.787 \pm 0.018$ & $0.838 \pm 0.018$ & $0.765 \pm 0.017$ & $0.834 \pm 0.031$ \\
\textbf{\coauthor} & $0.935 \pm 0.005$ & $0.939 \pm 0.005$ & $0.937 \pm 0.006$ & $0.940 \pm 0.003$ \\
\textbf{\dblp} & $0.706 \pm 0.010$ & $0.706 \pm 0.021$ & $0.708 \pm 0.011$ & $0.724 \pm 0.033$ \\
\textbf{\pubmed} & $0.832 \pm 0.010$ & $0.924 \pm 0.007$ & $0.841 \pm 0.007$ & $0.935 \pm 0.011$ \\
\hline
\end{tabular}
\label{tab:fine_tuning_performance}
\end{table*}

\noindent\textbf{Double extraction} involves \respadv running two model extraction attacks to obtain the final \surrogate: first against \target to get an intermediate model, followed by another attack against the intermediate model to obtain \surrogate. This additional attack is to make the \surrogate distinct from \target. We refer to such surrogates as \doublesurr

To satisfy the robustness experiment, \method should achieve low FNR against double extraction even if \doublesurr accuracy drops by up to 5\% points. An accuracy drop greater than 5\% points greatly reduces model utility. We use the previously trained \similarity directly and build additional \doublesurr models on the previously trained \target models. We use the same attack for both extractions since the difference between the two attacks is the knowledge of \respadv, and it is unrealistic for \respadv to have differing knowledge when running two attacks sequentially. As before, we use two architectures for each \doublesurr per \target, and create nine versions using random initialization (a total of 18 test models per \target). We repeat each experiment five times.

We evaluated \method against \doublesurr models reported in Table~\ref{tab:double_extraction_performance}. We find that while the adversary experiences a significant loss in utility (at least 3\% points in accuracy) on \acm, \amazon, and \citseer datasets, \method still achieves zero false negatives across all datasets. This shows that \method is effective in mitigating double extraction attacks across different datasets.

\begin{table*}[!htb]
\centering
\caption{Performance of \doublesurr. The model utility is comparable to \surrogate in most cases, but drops by $\approx8\%$ points for \amazon and $\approx12\%$ points for \citseer.}
\begin{tabular}{|c|c|c|c|c|}
\hline
\textbf{Dataset} & \textbf{Type 1} \doublesurr \textbf{Accuracy} & \textbf{Type 1 }\doublesurr \textbf{Fidelity} & \textbf{Type 2} \doublesurr \textbf{Accuracy} & \textbf{Type 2} \doublesurr \textbf{Fidelity} \\
\hline
\textbf{\acm} & $0.876 \pm 0.023$ & $0.932 \pm 0.018$ & $0.882 \pm 0.017$ & $0.930 \pm 0.020$ \\
\textbf{\amazon} & $0.780 \pm 0.046$ & $0.782 \pm 0.056$ & $0.698 \pm 0.216$ & $0.695 \pm 0.219$ \\
\textbf{\citseer} & $0.681 \pm 0.086$ & $0.719 \pm 0.086$ & $0.679 \pm 0.064$ & $0.736 \pm 0.093$ \\
\textbf{\coauthor} & $0.918 \pm 0.006$ & $0.943 \pm 0.011$ & $0.916 \pm 0.009$ & $0.943 \pm 0.004$ \\
\textbf{\dblp} & $0.686 \pm 0.009$ & $0.786 \pm 0.027$ & $0.678 \pm 0.019$ & $0.784 \pm 0.036$ \\
\textbf{\pubmed} & $0.829 \pm 0.005$ & $0.923 \pm 0.007$ & $0.831 \pm 0.004$ & $0.930 \pm 0.005$ \\
\hline
\end{tabular}
\label{tab:double_extraction_performance}
\end{table*}

\noindent\textbf{Pruning} removes model weights to reduce computational complexity while maintaining model utility. This alters the output distribution (in our case, embeddings), which could potentially affect the success of \method. 

We experiment with \emph{prune ratios} (ratio of all model weights set to 0) ranging from 0.1 to 0.7 as pruning beyond resulted in a high accuracy loss (>20\% points for all datasets). As before, \method is robust if it achieves low FNR against \surrogate models with less than a 5\% point drop in accuracy. We use the same experimental setup of training nine versions of \surrogate and pruning each one with the ratios above (a total of 18 test models per prune ratio).

\begin{figure}
    \centering
    \includegraphics[width=0.5\textwidth]{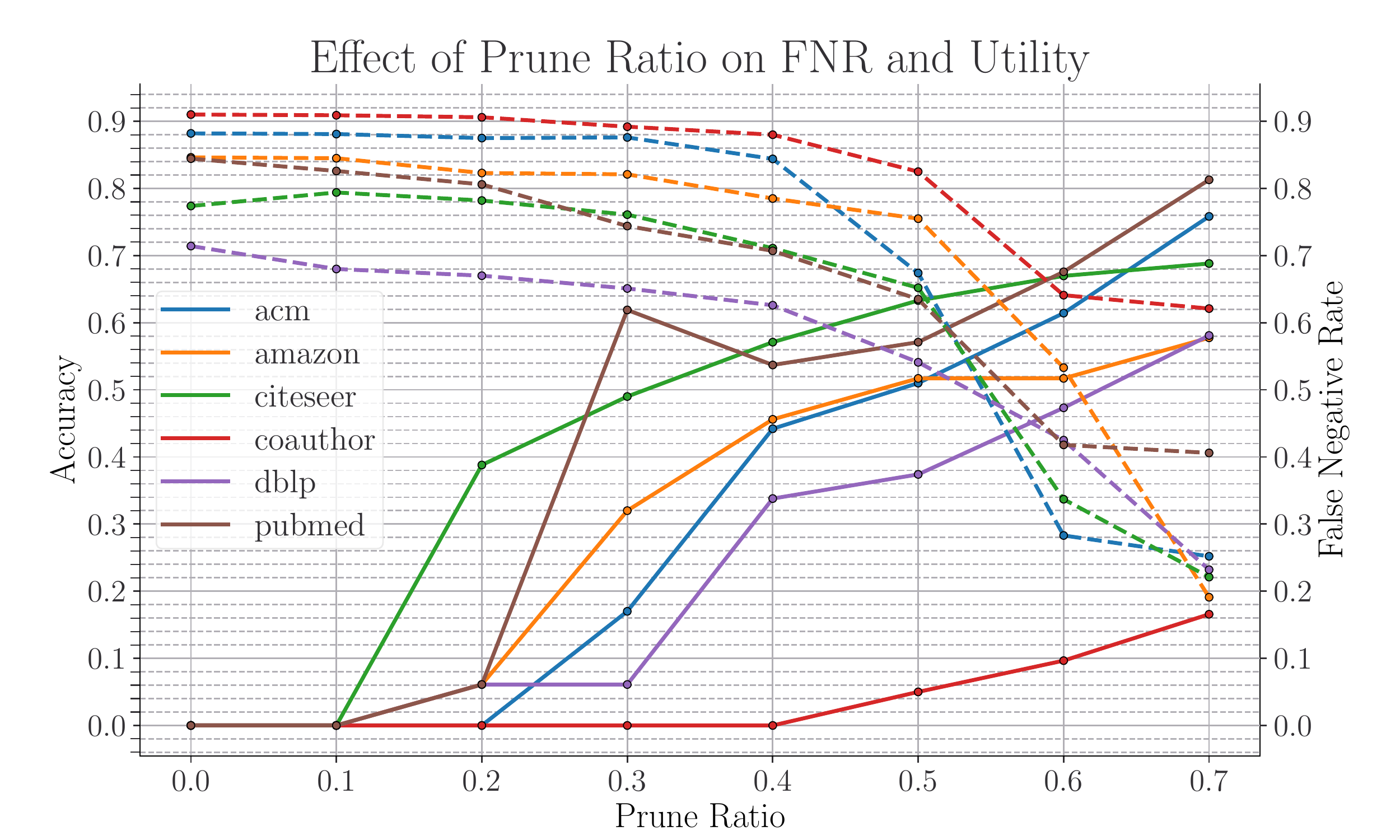}
    \caption{\method performance against pruning. Dotted lines represent \surrogate accuracy, and solid lines represent FNR. As the pruning ratio increases, the accuracy decreases, and the FNR increases. By default, \method fails against prune ratios of 0.3-0.4 for most datasets.}
    \label{fig:pruning_simple}
\end{figure}

We report our results in Figure~\ref{fig:pruning_simple}. We observe that \surrogate accuracy falls significantly after 0.4 (>5\% points for all datasets). Only the FNR for \coauthor stays small until a ratio of 0.6. We conjecture this is due to \coauthor being the largest dataset, it has 1.8 times more nodes and 5 times more edges than the second largest dataset. For the rest of the datasets the FNR increases around a ratio of 0.2 to 0.3. This shows that \method in its basic form fails against pruning.

\noindent\textbf{\method can be made robust} against pruning attacks by adversarially optimizing \similarity (Section~\ref{sec:approach}). We do this by including the output from pruned models up to a ratio of 0.4 into the training data. Beyond a prune ratio of 0.4, the accuracy drop is large enough (> 5\% points) to deter \respadv. Using the same experimental setup, we evaluate the success of the more robust \method to mitigate pruning.

\begin{figure}
    \centering
    \includegraphics[width=0.5\textwidth]{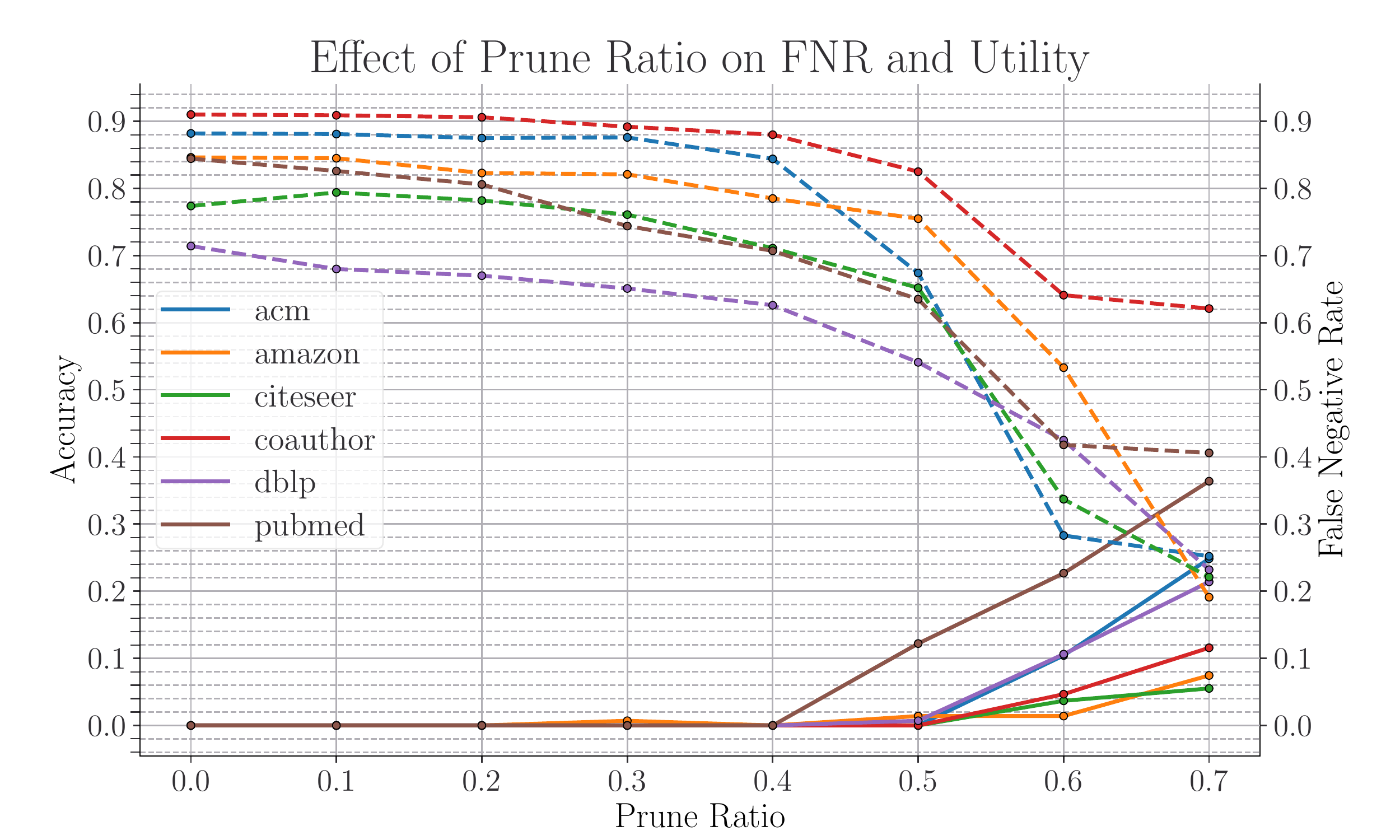}
    \caption{A more robust \method performs better against pruning. Dotted lines represent \surrogate accuracy, and solid lines represent FNR. Robust-\method works up till a prune ratio of 0.4, after which \surrogate utility decreases more than 5\%.}
    \label{fig:pruning_robust}
\end{figure}

We observe that \method still achieves zero false negatives against the basic and double extraction attacks mentioned before. Additionally, it achieves a lower FPR, with only a 1.4\% FPR for \amazon, 0.7\% FPR for \pubmed, and a 0\% FPR for the other datasets. We report the results for pruning in Figure~\ref{fig:pruning_robust}. It achieves nearly zero FNR for all datasets up to a prune ratio of 0.4. Noting the success of robust optimization, we conjecture that this can be used to make \method robust against future evasion techniques.

We, therefore, conclude that \method remains effective (\ref{req2}) even in the presence of adversaries (\ref{req3}). Furthermore, adversarially training \method allows \verifier to choose between resource utilization and performance. The larger and more diverse the training data for \similarity, the longer it would take to train \method, and the more robust it would be. We have shown the minimum training data required to perform well against the evasion techniques we identified. 

\noindent\textbf{Distribution-shift} is a method that allows \respadv to adversarially train \surrogate to change the output distribution to make the embeddings distinct from \target's embeddings. This is constrained to minimize degradation of \surrogate accuracy.

We designed an experiment to reproduce this by training \surrogate with an adversarial autoencoder that shifts the output distribution of \surrogate to a Gaussian distribution. In this, \surrogate is treated as a generator and is trained simultaneously with a discriminator that detects the Gaussian distribution. We could not successfully modify the output distribution of \surrogate, and \method could successfully classify all such models with zero false negatives. We conjecture that the high-fidelity requirement forces the output distribution of \surrogate to be similar to \target.

\subsection{Efficiency of \method}\label{eval:efficiency}
\begin{topics}
    \topic{\similarity takes an average of XX seconds to train. Once trained, inference time is only XX seconds.}
\end{topics}

\begin{table*}[h]
\caption{Average total time over five runs in seconds (with 95\% confidence intervals) taken to train \target, and generate training data and train \similarity.}
\centering
\begin{tabular}{|c|c|c|c|c|c|c|}
\hline
& \multicolumn{6}{c|}{\textbf{Architectures}}\\
& \multicolumn{2}{c|}{\textbf{GAT}} & \multicolumn{2}{c|}{\textbf{GIN}} & \multicolumn{2}{c|}{\textbf{GraphSAGE}} \\
\textbf{Dataset} & \target & \similarity & \target & \similarity & \target & \similarity \\
\hline
\textbf{\coauthor} & $1184 \pm 53$ & $10562 \pm 1548$ & $1060 \pm 55$ & $10668 \pm 1205$ & $855 \pm 34$ & $10550 \pm 1237$ \\
\textbf{\pubmed} & $435 \pm 25$ & $3961 \pm 492$ & $418 \pm 26$ & $3845 \pm 257$ & $374 \pm 25$ & $3856 \pm 288$ \\
\textbf{\dblp} & $459 \pm 30$ & $4182 \pm 462$ & $412 \pm 26$ & $4011 \pm 498$ & $397 \pm 25$ & $3730 \pm 202$ \\
\textbf{\amazon} & $379 \pm 26$ & $3312 \pm 218$ & $361 \pm 25$ & $3273 \pm 171$ & $348 \pm 21$ & $3473 \pm 323$ \\
\textbf{\citseer} & $389 \pm 19$ & $3312 \pm 124$ & $357 \pm 24$ & $3142 \pm 204$ & $349 \pm 29$ & $2970 \pm 186$ \\
\textbf{\acm} & $334 \pm 27$ & $2985 \pm 165$ & $343 \pm 27$ & $2943 \pm 223$ & $351 \pm 33$ & $2876 \pm 134$ \\
\hline
\end{tabular}
\label{tab:efficiency}
\end{table*}

We now evaluate the efficiency of \method (requirement \ref{req4}). We want to ensure the computation overhead of \method is reasonable.

To this end, we measure the execution time to generate the training data for \similarity and train it. Recall from Section~\ref{setup:architectures} that we train multiple versions of \independent and \surrogate to generate the training data for \similarity. For \method, this involves building six versions of \independent~(two models per architecture with different random initializations) and two versions of \surrogate along with their pruned variants (ten versions of \surrogate). These models can be trained in parallel, making it faster. However, we train them sequentially to accurately measure the total time taken, assuming just one available machine. The models were trained on a machine with two AMD EPYC 7302 16-Core CPUs and eight Nvidia A100 GPUs with 40GB VRAM per GPU. 

We summarize the results for each target model in Table~\ref{tab:efficiency}. 
It generally takes less than 3 hours to train \method. The time taken to train \method includes the time taken to train the additional surrogate and independent models to generate the training data and to train \similarity. The size of the dataset influences the time taken to train the additional models. \coauthor is the largest dataset, with the longest training times, followed by \dblp and \pubmed. The variation in time taken within datasets is caused by the optimization of \similarity. For instance, for \amazon, the time taken to train \similarity for GAT-based \target was 187 seconds on average, while for GIN-based \target it took 340 seconds on average. There is no trend for which architecture takes the longest time; it is affected by the combination of dataset and architecture. \method built for \acm-based models was the fastest to train, with \similarity training taking less than 100 seconds. For \coauthor, however, \similarity training took much longer, between 1561 to 2103 seconds.

Recall that in the practical setting we describe in Section~\ref{prob:sysmodel}, \verifier only trains \method when \accuser initiates a dispute with \responder. Considering most models will not encounter ownership disputes, we consider the numbers in Table~\ref{tab:efficiency} to be reasonable. 

In conclusion, we show that \method satisfies the requirement of being an effective (\ref{req2}) robust (\ref{req3}) and efficient (\ref{req4}) fingerprinting scheme. 
\section{Related Work}\label{sec:related}

In addition to the prior work directly related to \method (Section~\ref{back:modelext}), we broadly discuss other related work pertaining to privacy and security in GNNs.

\noindent\textbf{\underline{Robustness Attacks against GNNs}} find adversarial examples modifying the adjacency matrix to perturb the input, which results in misclassification~\cite{wu2019Adversarial, dai2018AdversarialAttack, entezari2020Adversarial, zugner2018Adversarial, zugner2020Adversarial}. As mentioned in Section~\ref{prob:limitations}, finding adversarial examples for graphs is not trivial. The best attack rate in these prior works is 35\%. For comparison, image-based robustness attacks can reach nearly a 100\% attack success rate~\cite{dong2018adversarialImage}.

\noindent\textbf{\underline{Defences against Robustness Attacks}} limit the effect of perturbed edges on the GNN~\cite{zhang2020RobustGNNGuard, deng2022RobustGNNGarnet, entezari2020Adversarial, miller2019RobustGNN, wu2019Adversarial, zhang2019RobustGNN} or adversarial training (training with adversarial examples) to make the model robust~\cite{jin2019RobustGNN, sun2019RobustGNN, tang2020RobustGNN}. 
Some defenses prune edges in the adjacency matrix to minimize the effect of the adversarial perturbations on edges with the intuition that homophily ensures nodes and edges connecting them are similar~\cite{wu2019Adversarial, zhang2020RobustGNNGuard}. Hence, any non-similar edges are potentially adversarial and can be pruned. This can also be extended to cases where homophily is not satisfied~\cite{deng2022RobustGNNGarnet}. RobustGCN~\cite{zhu2019RobustGNN} trains the model such that the embeddings follow a Gaussian distribution to reduce the effect of perturbed edges.

\noindent\textbf{\underline{Privacy Attacks against GNNs}} aim to infer sensitive unobservable information pertaining to training dataset and inputs given access to the model. 
Membership inference attacks infer whether a node/sub-graph was used to train the model\cite{olatunji2021NodeLevel, he2021NodeLevel, duddu2020quantifying, wu2021GraphInference}. Olatunji et al.~\cite{olatunji2021NodeLevel} further present a node-level membership inference attack using the 2-hop subgraph of the target node. He et al.~\cite{he2021NodeLevel} achieve a similar goal, except they use only the 0-hop subgraph. Wu et al.~\cite{wu2021GraphInference} present the first work on graph-level membership inference attack. 
Access to the embeddings can also allow adversaries to reconstruct the training graph~\cite{duddu2020quantifying}.
Link inference attacks identify whether there exists an edge between two query nodes\cite{he2021StealingLinks,zhang2022InferenceAttacks}. 
Property inference attacks infer global properties about the training graph datasets (e.g., average clustering coefficient, the proportion of males and females)~\cite{suri2021PropertyInference, suri2023PropertyInference, wang2022PropertyInference}.

\section{Discussion and Conclusions}\label{sec:discussions}
\begin{topics}
    \topic{There are evasion techniques that are yet to be studied in detail. One potential technique can include an adversarial training mechanism to change output distribution. Our experimentation showed that this technique was ineffective. The goal of high-fidelity conflicts with adversarial training.}
    \topic{\method may fail against attacks that do not try to optimize fidelity between \target and \surrogate. However, there are no such attacks in the current literature. Such attacks do not transfer the behavior of the model exactly. Thus, they cannot be used to study \target other kinds of attacks, such as adversarial attacks.}
\end{topics}

We summarize our contributions and discuss some implications of our system model assumptions and design choices as well as potential limitations of \method. 
\subsection{Summary}\label{sec:summary}
In this paper, we proposed \method, a technique for fingerprinting GNNs and showed that it is effective against the state-of-the-art model extraction attack against GNNs described by Shen et al.~\cite{shen2022GNNInductiveSteal}. Further, we showed that \method is robust against various ways in which an adversary can try to alter the fingerprint of the surrogate model.

\subsection{Low-Fidelity Model Extraction Attacks}\label{sec:lowfid}

Recall that Shen et al.'s~\cite{shen2022GNNInductiveSteal} attack aims to maximize fidelity between \target and \surrogate. Consequently, \method relies on the fact that the embeddings of \target and \surrogate are similar. Low-fidelity extraction attacks will not necessarily result in \surrogate being close to \target, and can be potentially missed by \method. However, there are no low-fidelity GNN extraction attacks in the current literature.

\subsection{Requirement for Model Registration}\label{sec:modelreg}

Following prior work~\cite{szyller2021WatermarkingDawn,adi2018Backdooring,liu2023false}, in our system model (Section~\ref{prob:sysmodel}), all model owners are required to obtain secure certified timestamps for their models before deployment to protect against different types of adversarial behavior.
Model owners can be incentivized to do so because: 
\begin{itemize}[leftmargin=*]
    \item registration serves as a direct means of protecting models against theft,
    \item it may be necessary to meet regulatory requirements for deployed models (e.g., EU guidelines~\cite{artificialintelligenceactHome}),
    \item in the future, model insurance mechanisms may be used to mitigate the effects of model theft, and model registration may become a requirement for insurance.
\end{itemize}

\subsection{Revisiting \accadv}\label{sec:malacc}

In Section~\ref{prob:sysmodel}, we argued that \accadv is thwarted by the model registration requirement. An alternative approach to trigger false positives in model ownership schemes is to generate the fingerprints adversarially. In all prior model ownership schemes, (both fingerprinting and watermarking)~\cite{lukas2021Conferrable, peng2022FingerprintUAP}, this is potentially feasible since it is the \accuser who generates the fingerprint/watermark. This is not the case in \method: the fingerprint is chosen by \verifier, rendering \method immune to any such adversarial fingerprint generation attacks.

\subsection{Shen et al.'s~\cite{shen2022GNNInductiveSteal} Prediction-based Attack}\label{sec:predatt}

In our adversary model (Section~\ref{prob:advmodel}), we consider that \target outputs embeddings as done in Shen et al.~\cite{shen2022GNNInductiveSteal}. This allows us to use \method over the outputs of \target, making it a black-box scheme. However, Shen et al.'s~\cite{shen2022GNNInductiveSteal} also suggest a prediction-based model extraction attack for cases where \target only outputs classification labels.
In such a setting, \method is still applicable if \verifier is given white-box access to \target.

\section*{Acknowledgments}
The authors would like to thank the anonymous reviewers for their valuable feedback. This work was supported in part by Intel (in the context of
the Private-AI Institute).
\bibliographystyle{IEEEtran}
\bibliography{IEEEabrv,paper}
\appendices

\section{Dataset Ownership Graphs}\label{appendix:dataset_ownership}
\begin{figure}
    \centering
    \subfigure[Diff. Training Data - \acm]{\includegraphics[width=0.23\textwidth]{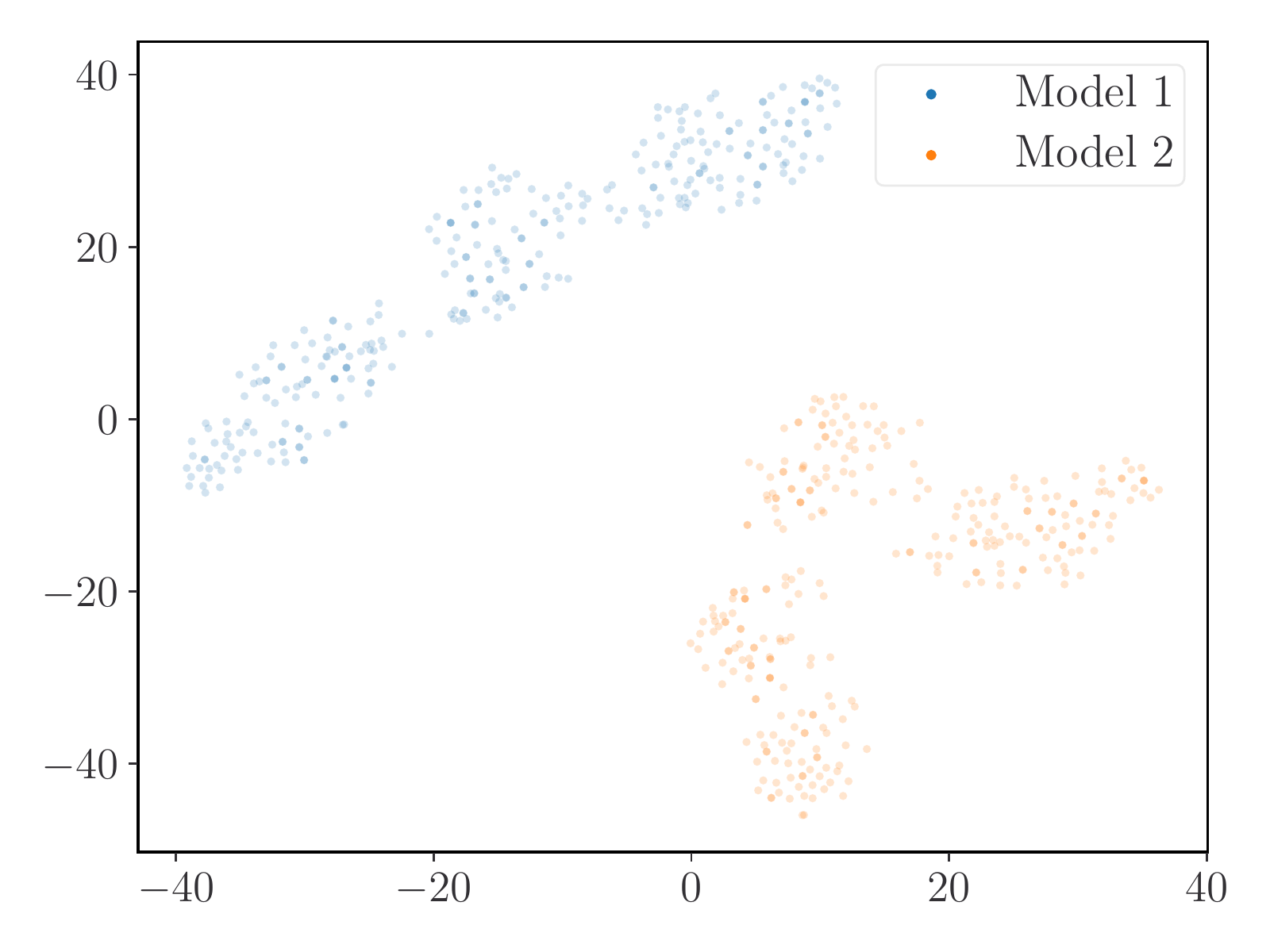}}
    \subfigure[Same Training Data - \acm]{\includegraphics[width=0.23\textwidth]{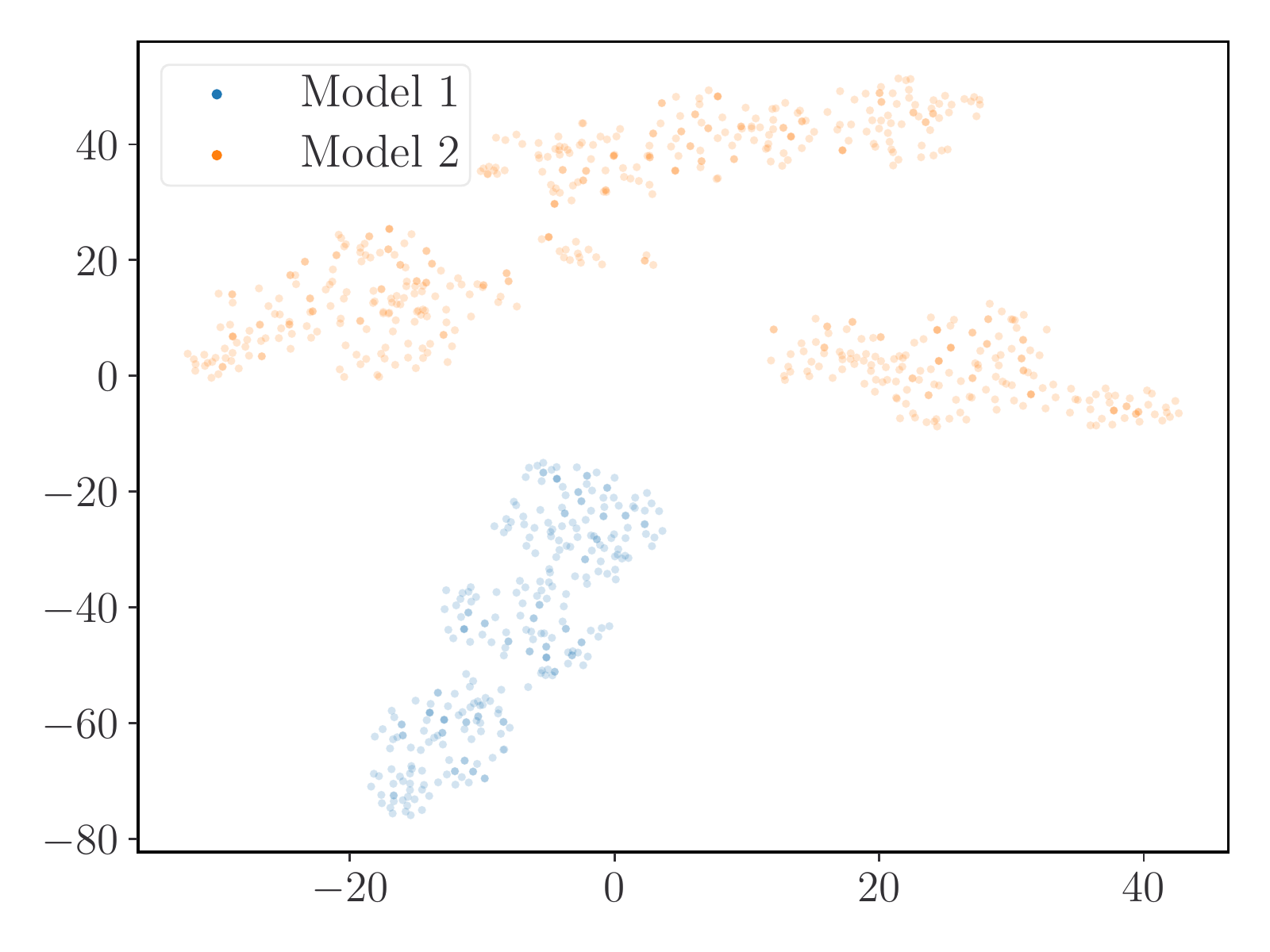}}
    \subfigure[Diff. Training Data - \amazon]{\includegraphics[width=0.23\textwidth]{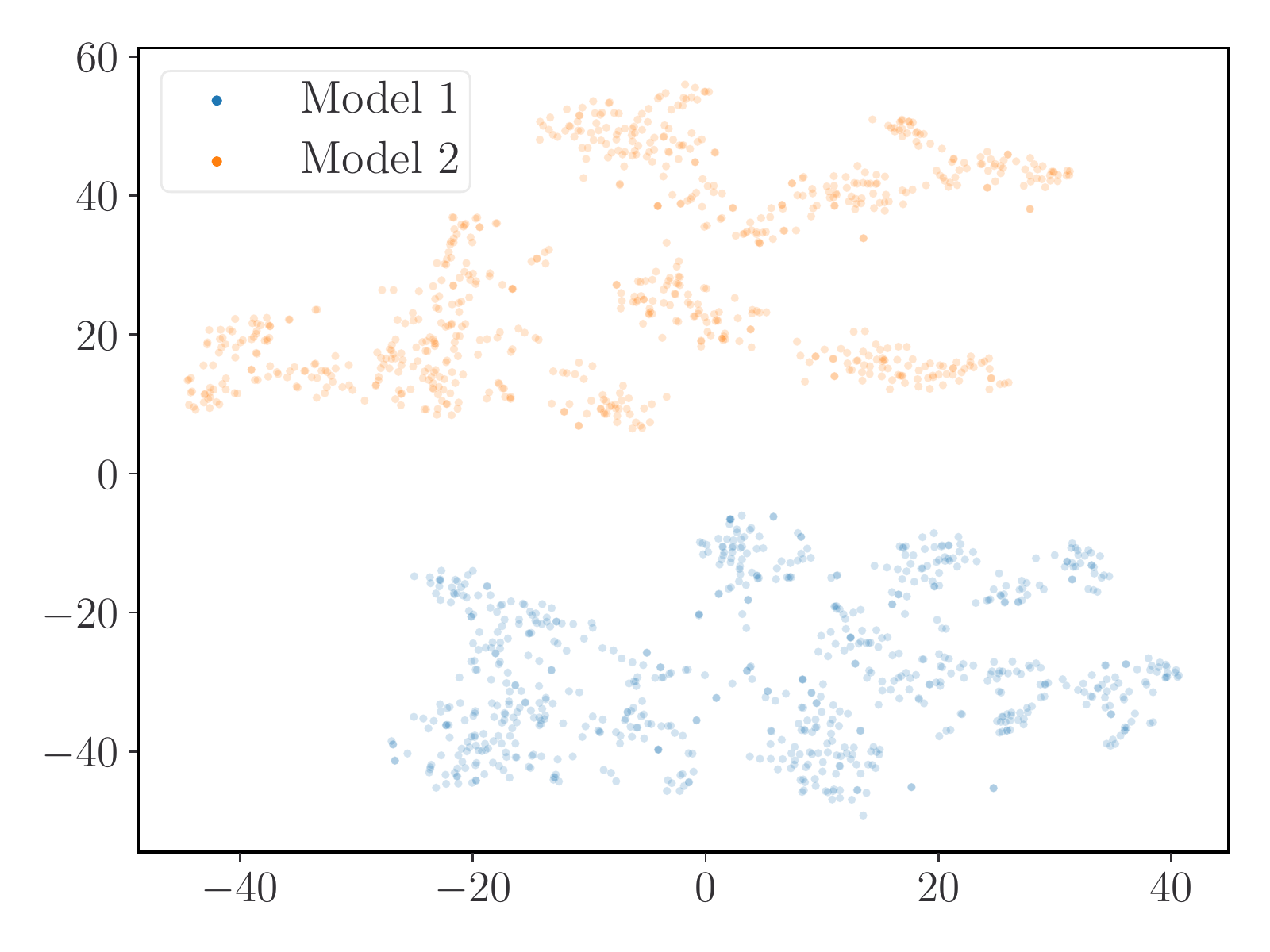}}
    \subfigure[Same Training Data - \amazon]{\includegraphics[width=0.23\textwidth]{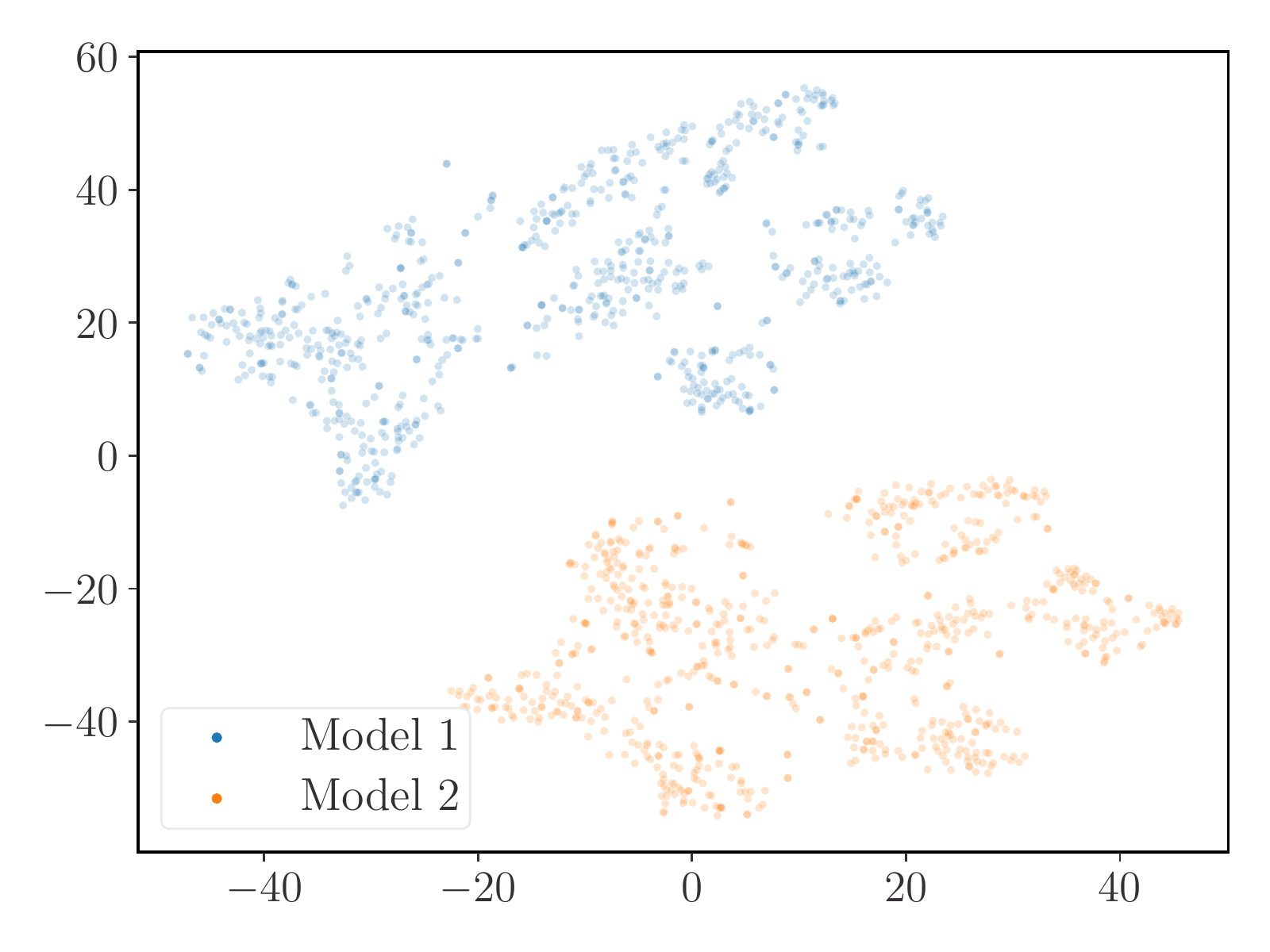}}
    \subfigure[Diff. Training Data - \dblp]{\includegraphics[width=0.23\textwidth]{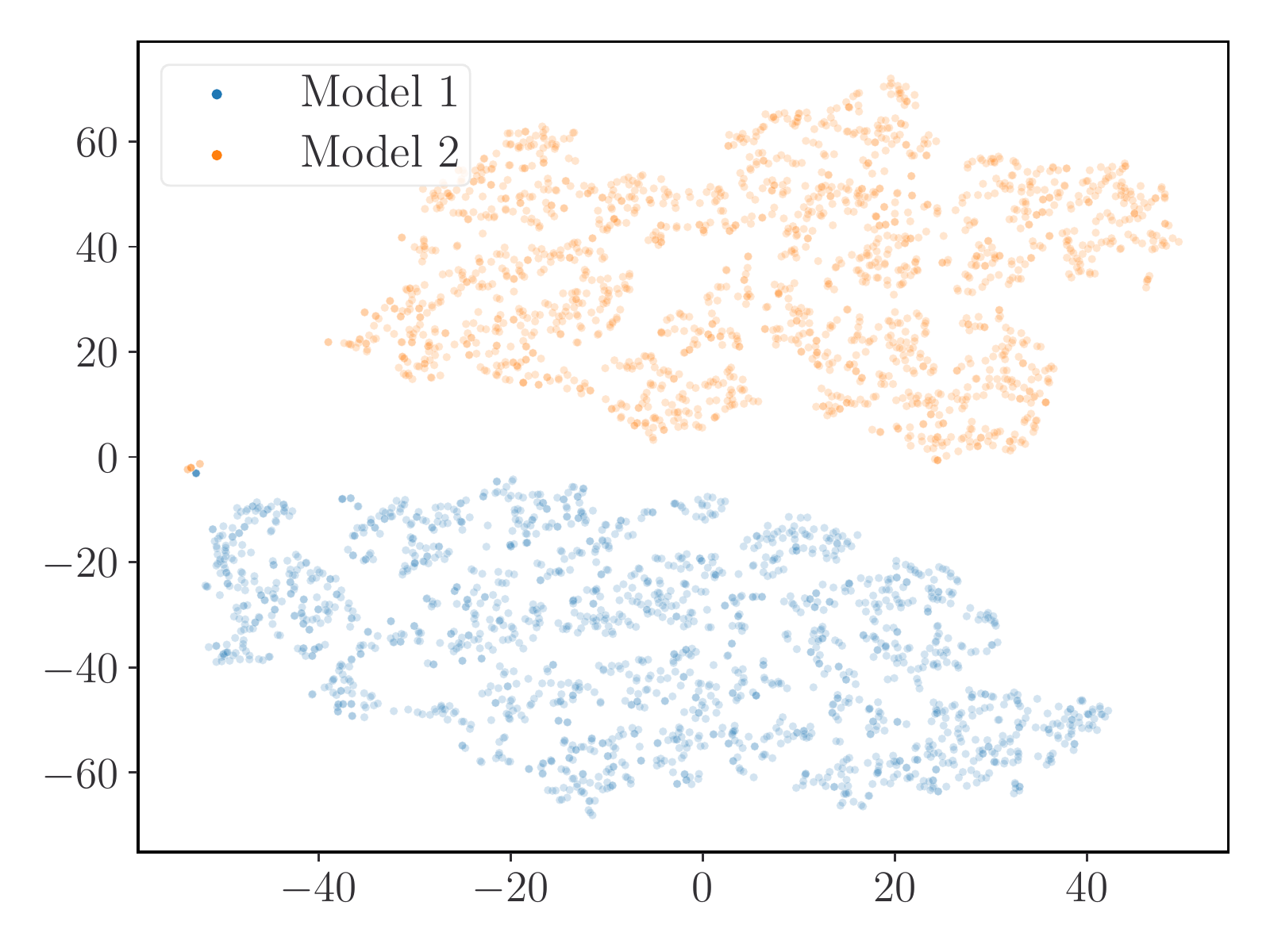}}
    \subfigure[Same Training Data - \dblp]{\includegraphics[width=0.23\textwidth]{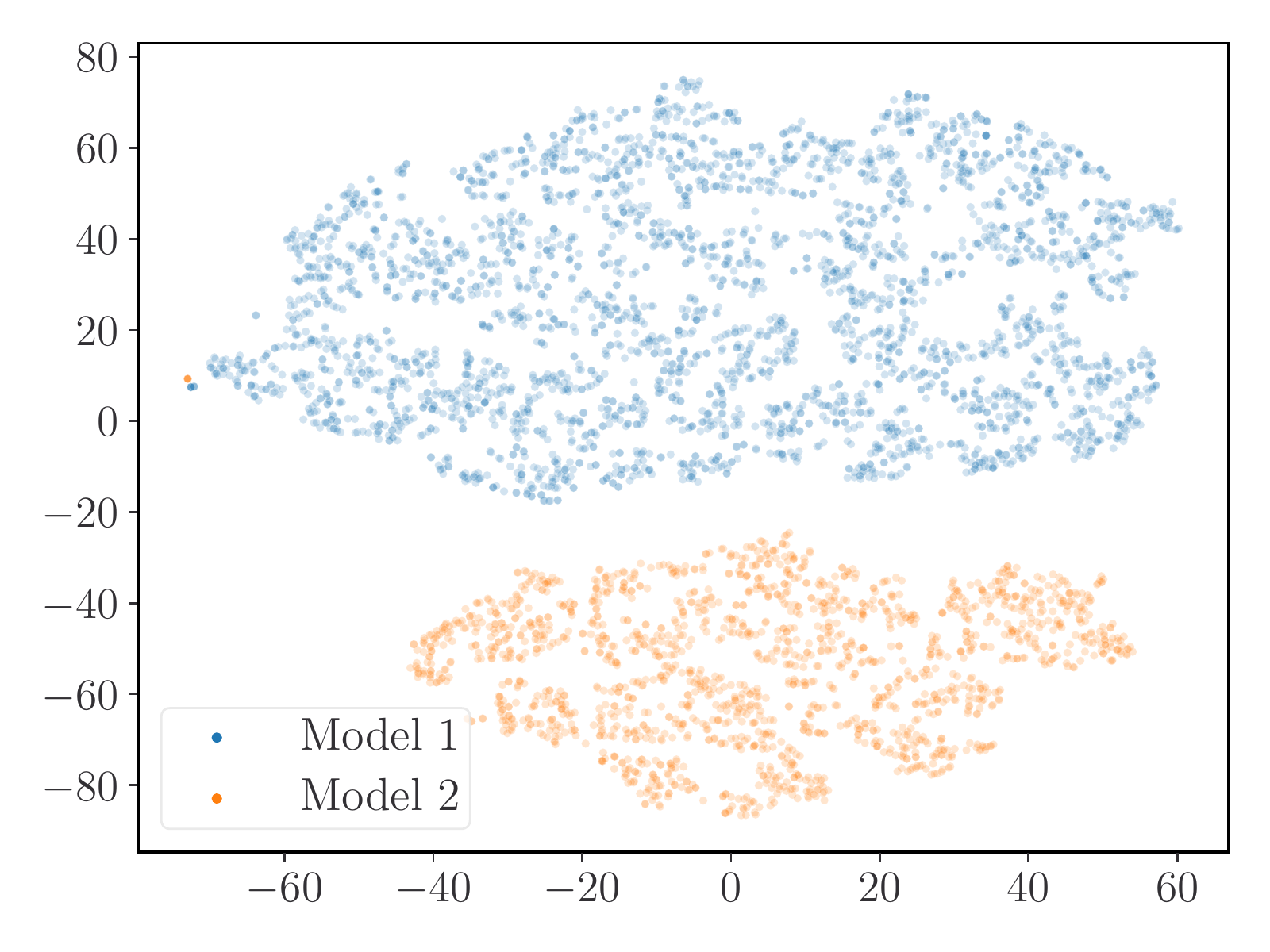}}
    \subfigure[Diff. Training Data - \citseer]{\includegraphics[width=0.23\textwidth]{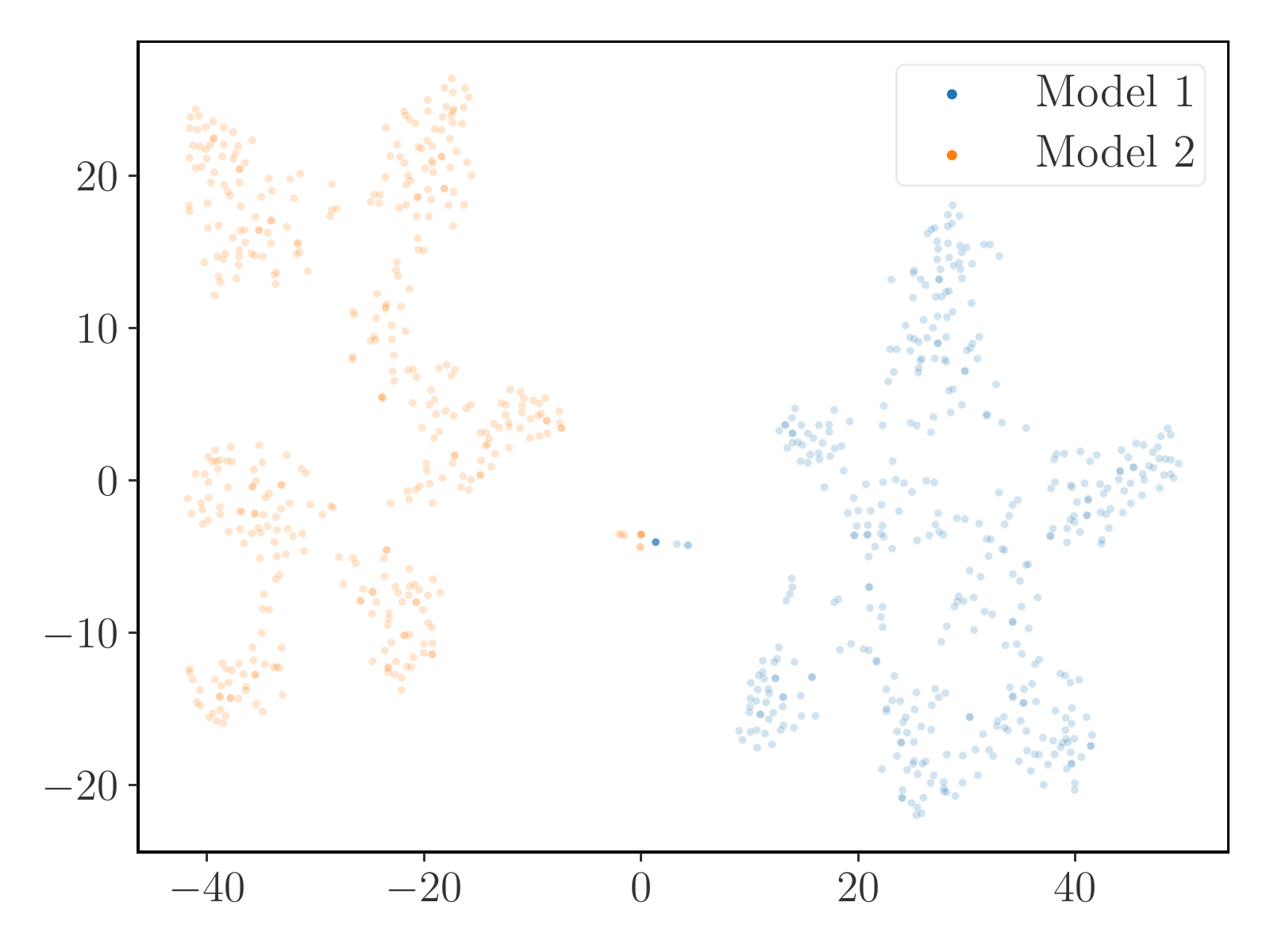}}
    \subfigure[Same Training Data - \citseer]{\includegraphics[width=0.23\textwidth]{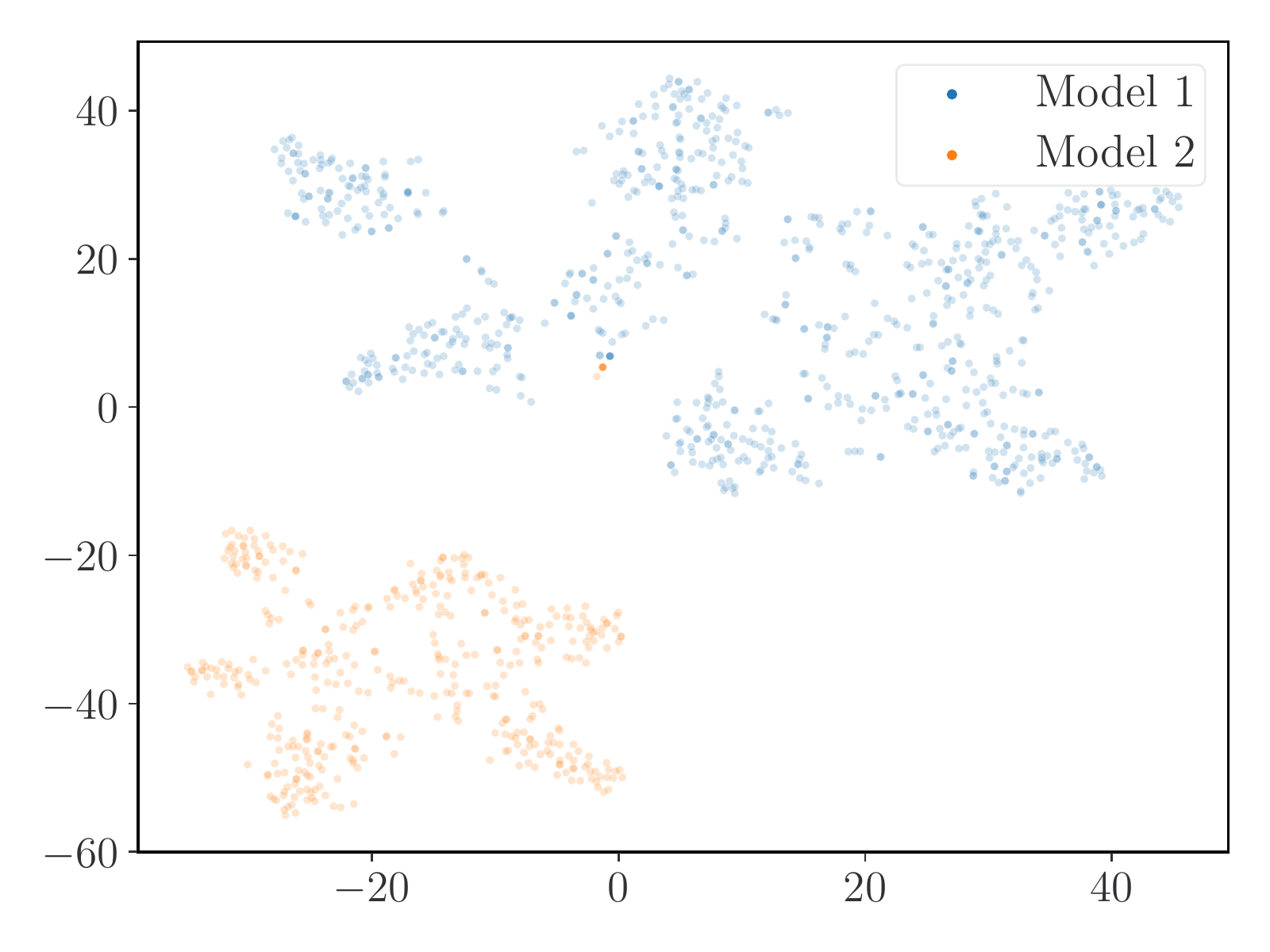}}
    \subfigure[Diff. Training Data - \pubmed]{\includegraphics[width=0.23\textwidth]{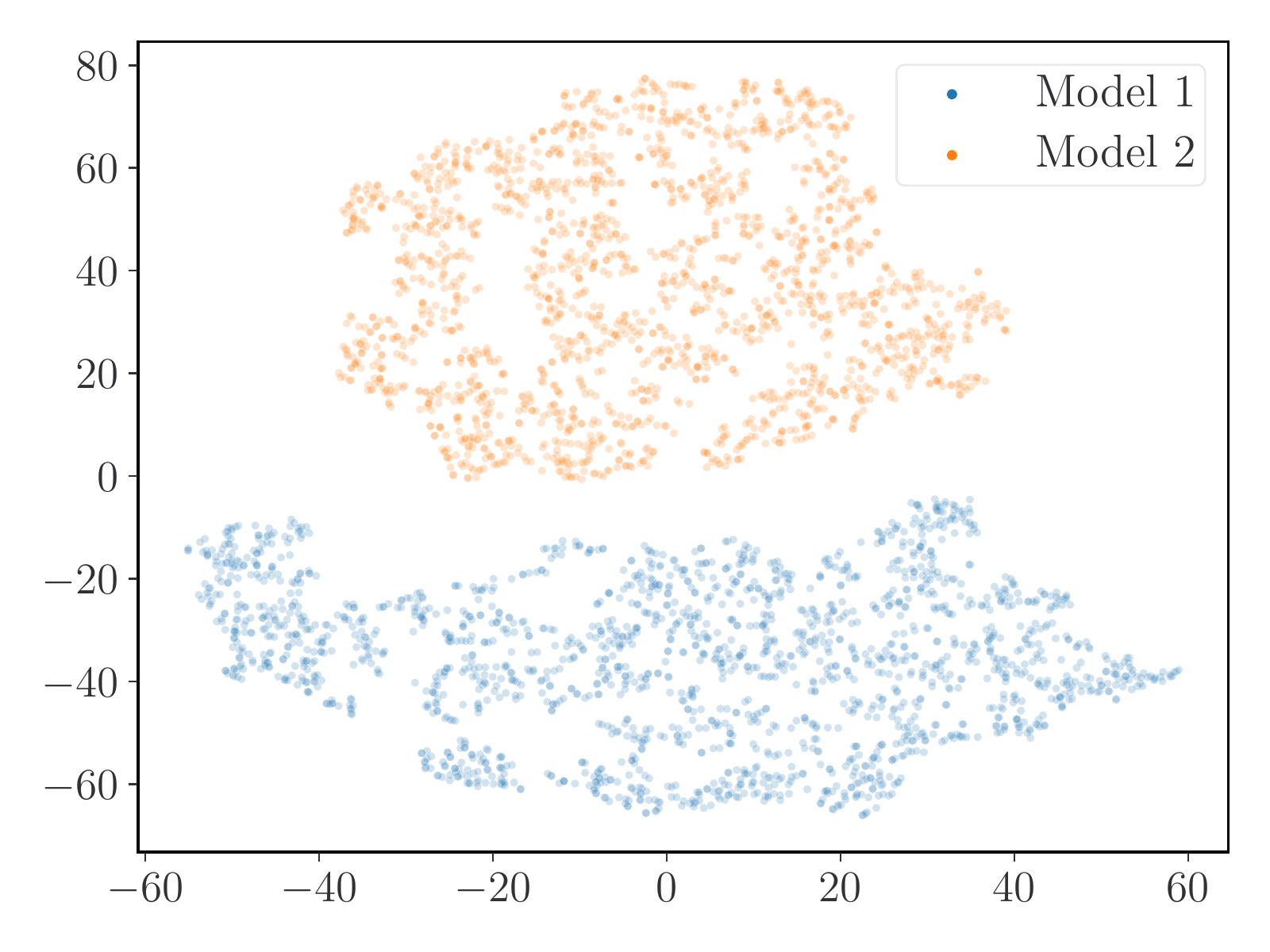}}
    \subfigure[Same Training Data - \pubmed]{\includegraphics[width=0.23\textwidth]{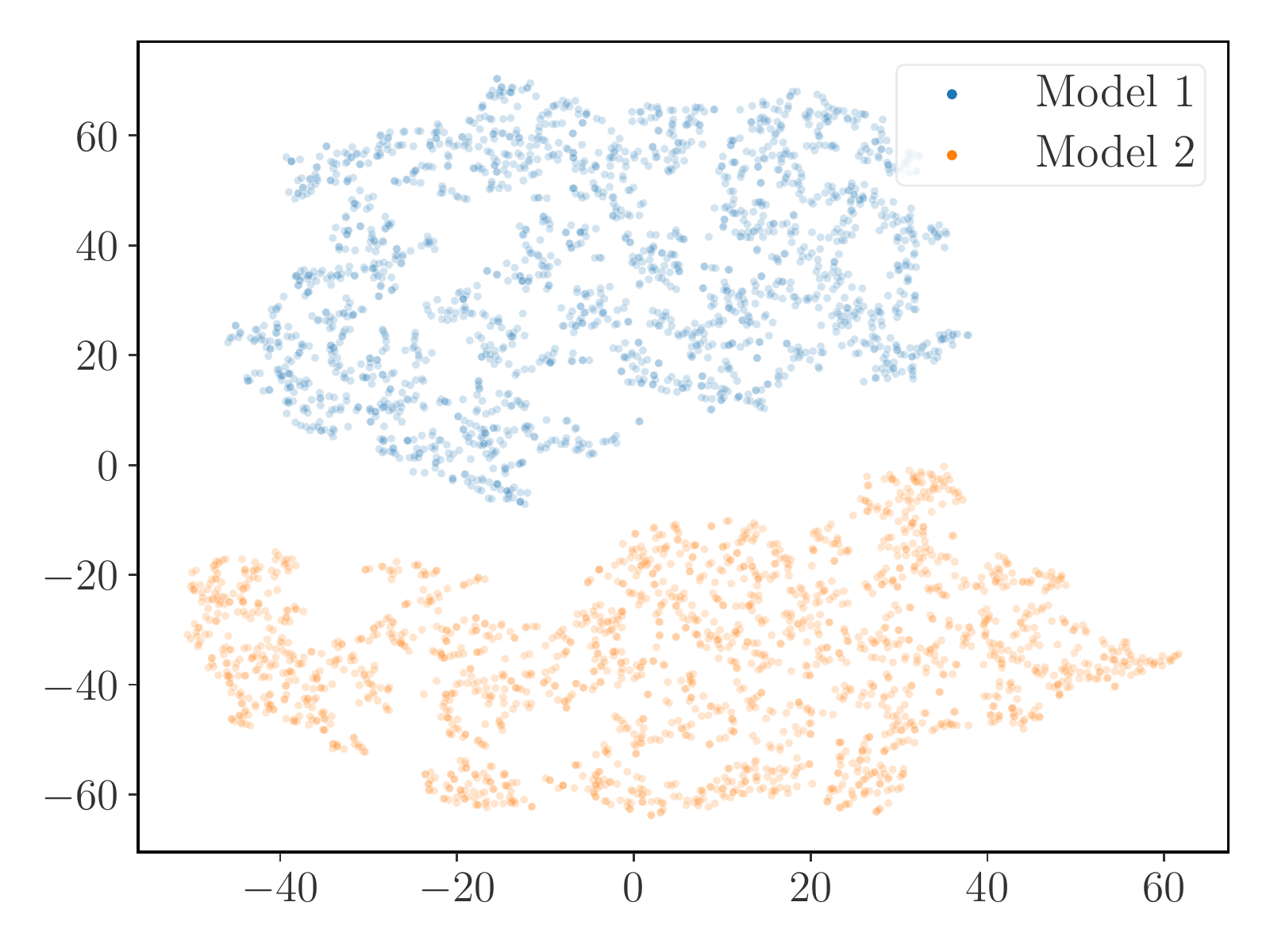}}
    \caption{t-SNE projections of the embeddings from two models trained are distinguishable for all datasets, regardless of whether the architecture is the same or different.}
    \label{appendix_fig:dataset_ownership}
\end{figure}

The difference in the two models is most easy to see using \coauthor since that is the largest dataset. However, the same trend is seen in the rest of the plots, as shown in Figure~\ref{appendix_fig:dataset_ownership}. Due to space constraints, we only show the graphs for models trained with GAT. The same trend is seen across all architectures.

\section{Model Ownership Graphs}\label{appendix:model_ownership}

\begin{figure}
    \centering
    \subfigure[Models trained on \acm dataset]{\includegraphics[width=0.36\textwidth]{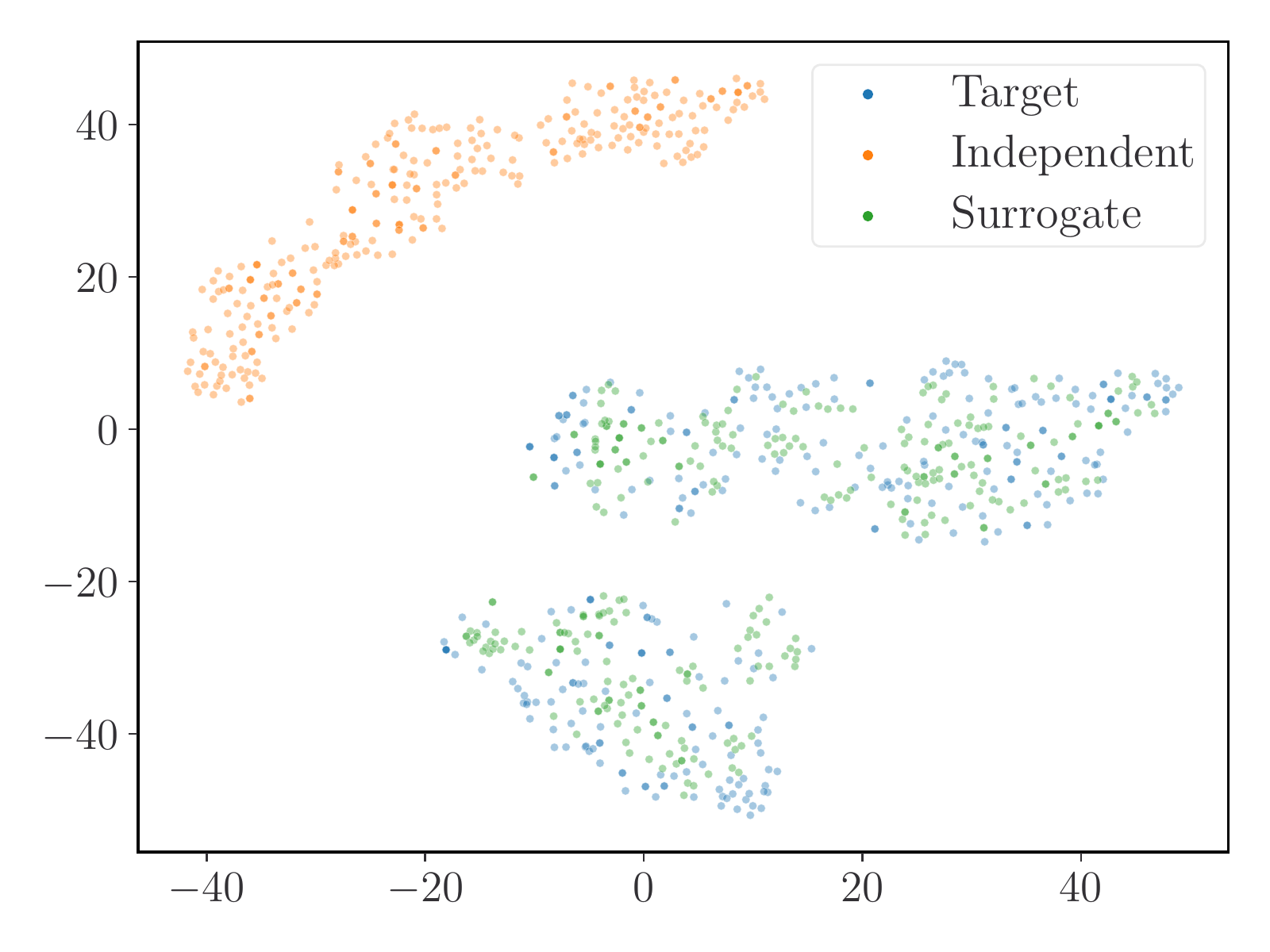}}
    \subfigure[Models trained on \amazon dataset]{\includegraphics[width=0.36\textwidth]{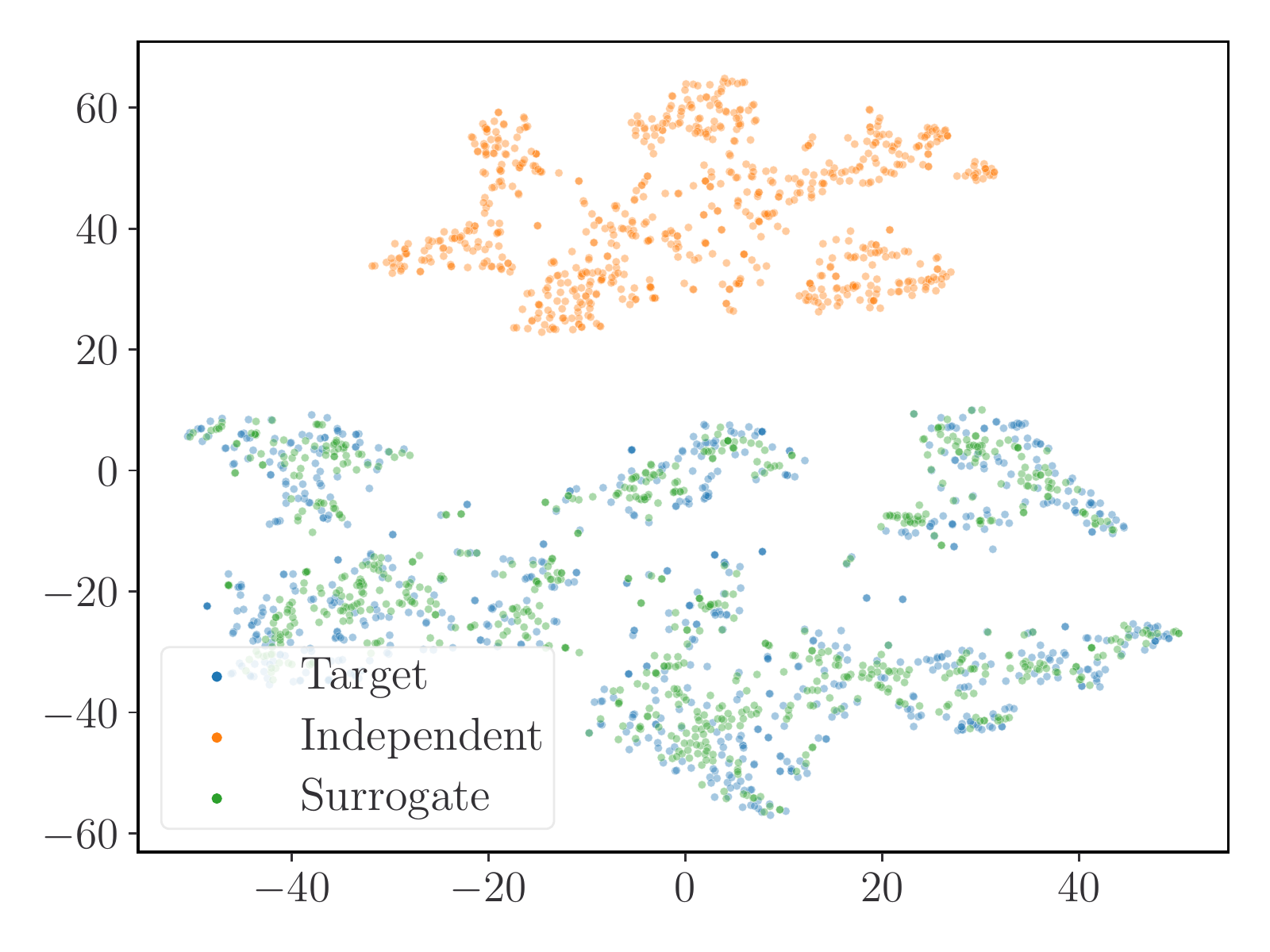}}
    \subfigure[Models trained on \citseer dataset]{\includegraphics[width=0.36\textwidth]{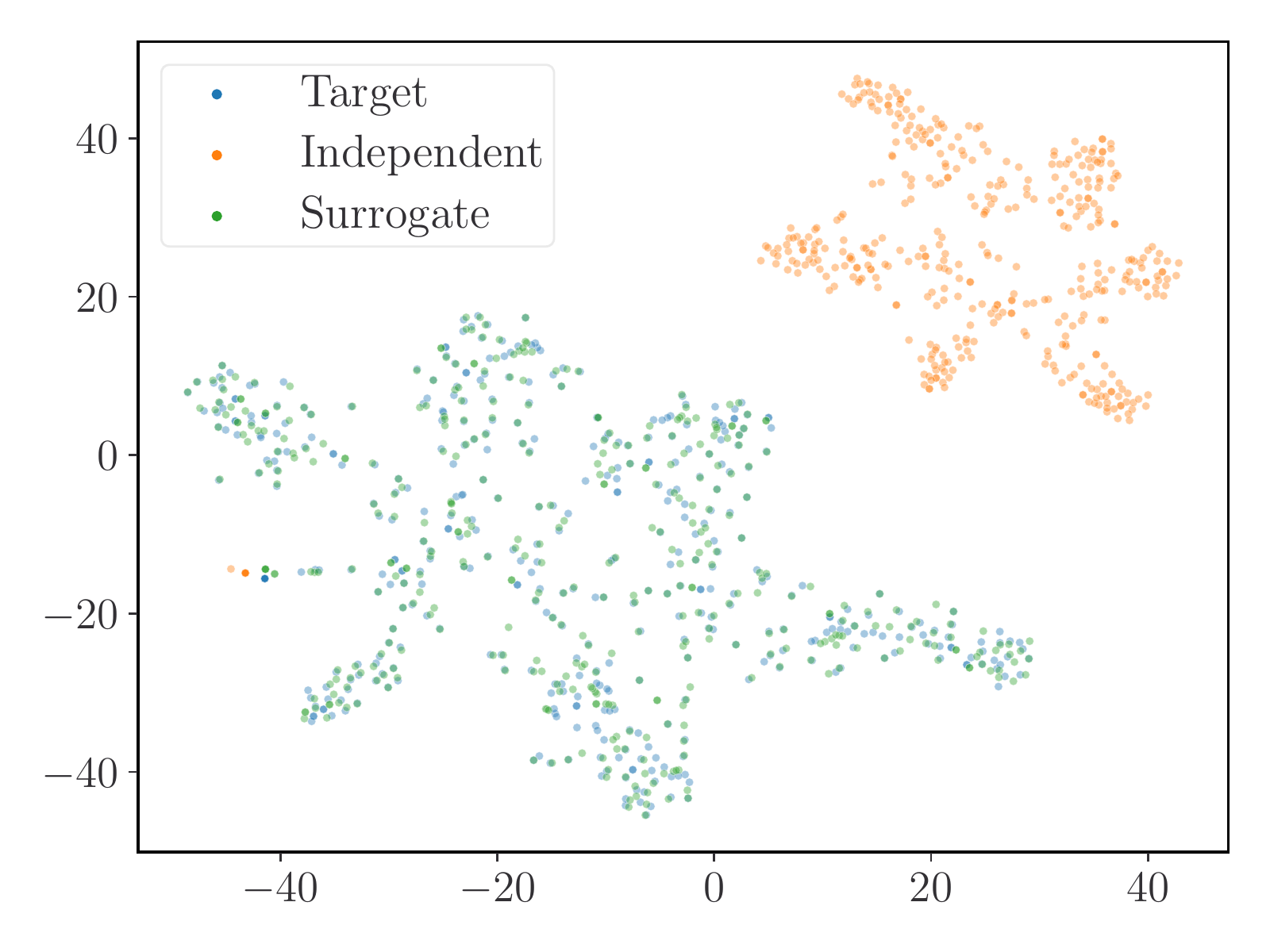}}
    \subfigure[Models trained on \pubmed dataset]{\includegraphics[width=0.36\textwidth]{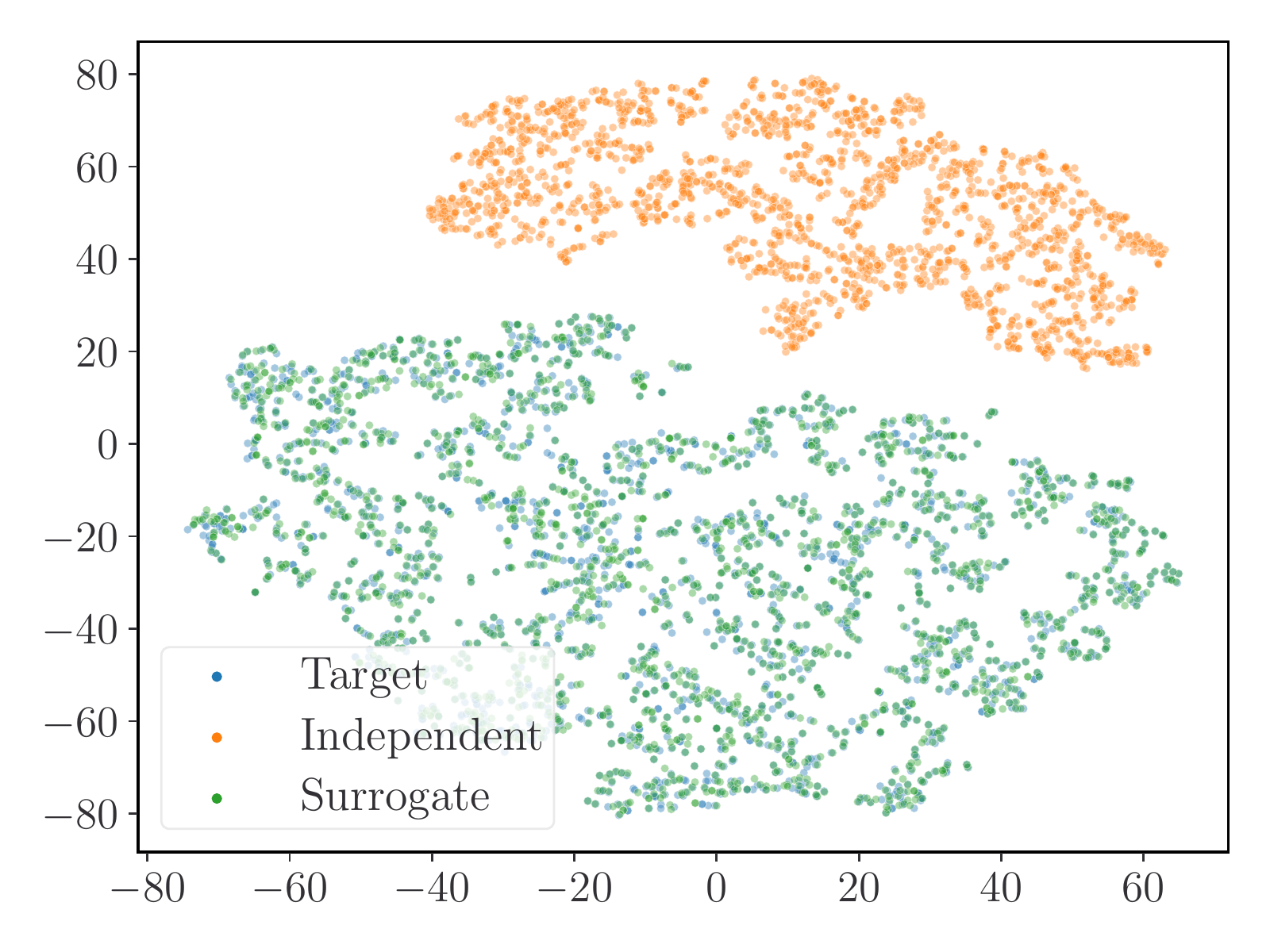}}
    \caption{t-SNE projections of the embeddings from \surrogate and \target overlap, while those from \independent are distinct. All models in this figure use the GAT architecture. The same trend holds for all architectures.}
    \label{appendix_fig:model_ownership}
\end{figure}

We show the graphs for the model ownership experiment in Figure~\ref{appendix_fig:model_ownership}. Due to space constraints, we only show the graphs for models trained with GAT. The same trend is seen across all architectures.

\section{Distance Graphs}\label{app:distances}
\begin{figure}
    \centering
    \subfigure[\acm]{\includegraphics[width=0.23\textwidth]{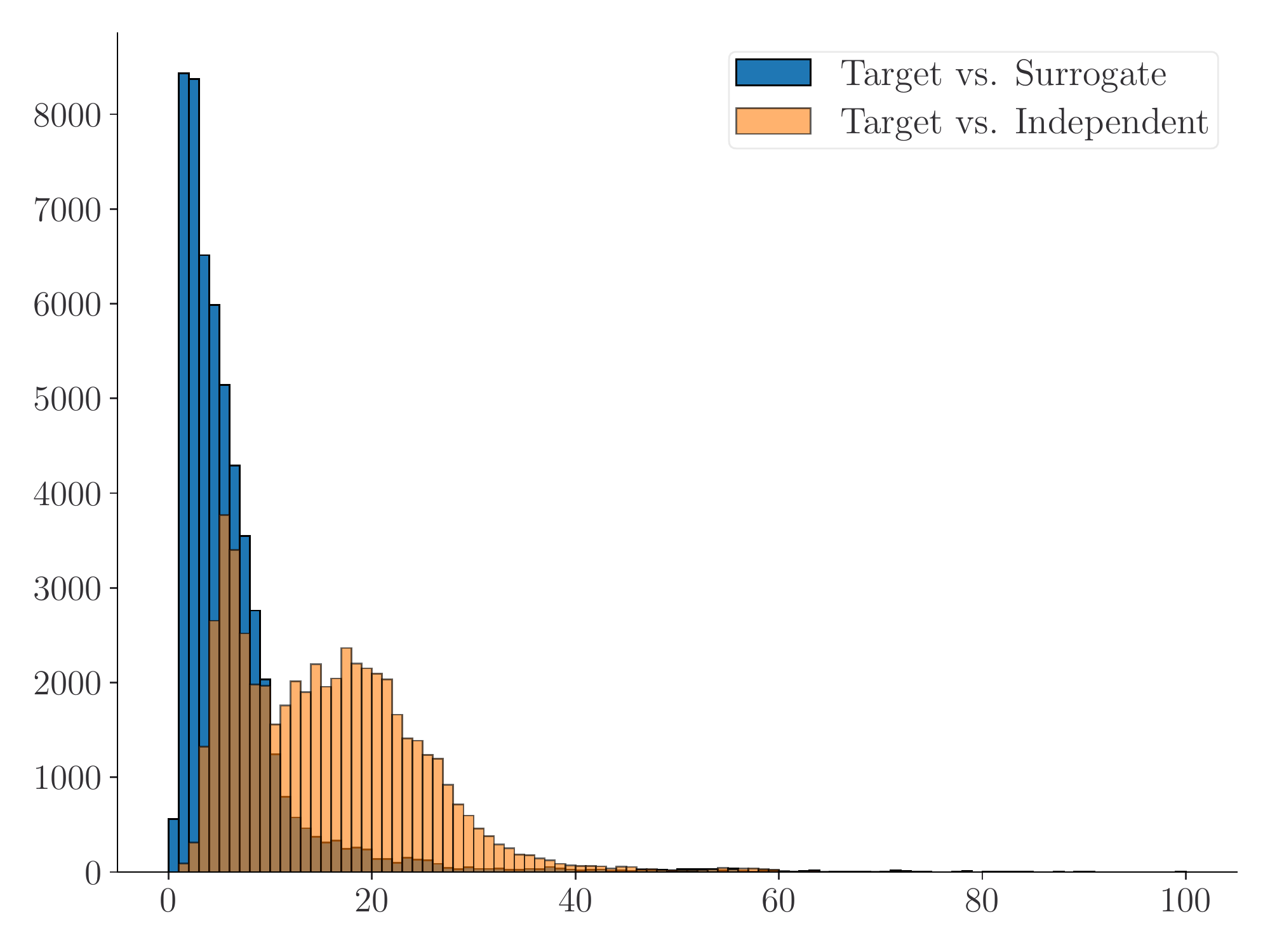}}
    \subfigure[\amazon]{\includegraphics[width=0.23\textwidth]{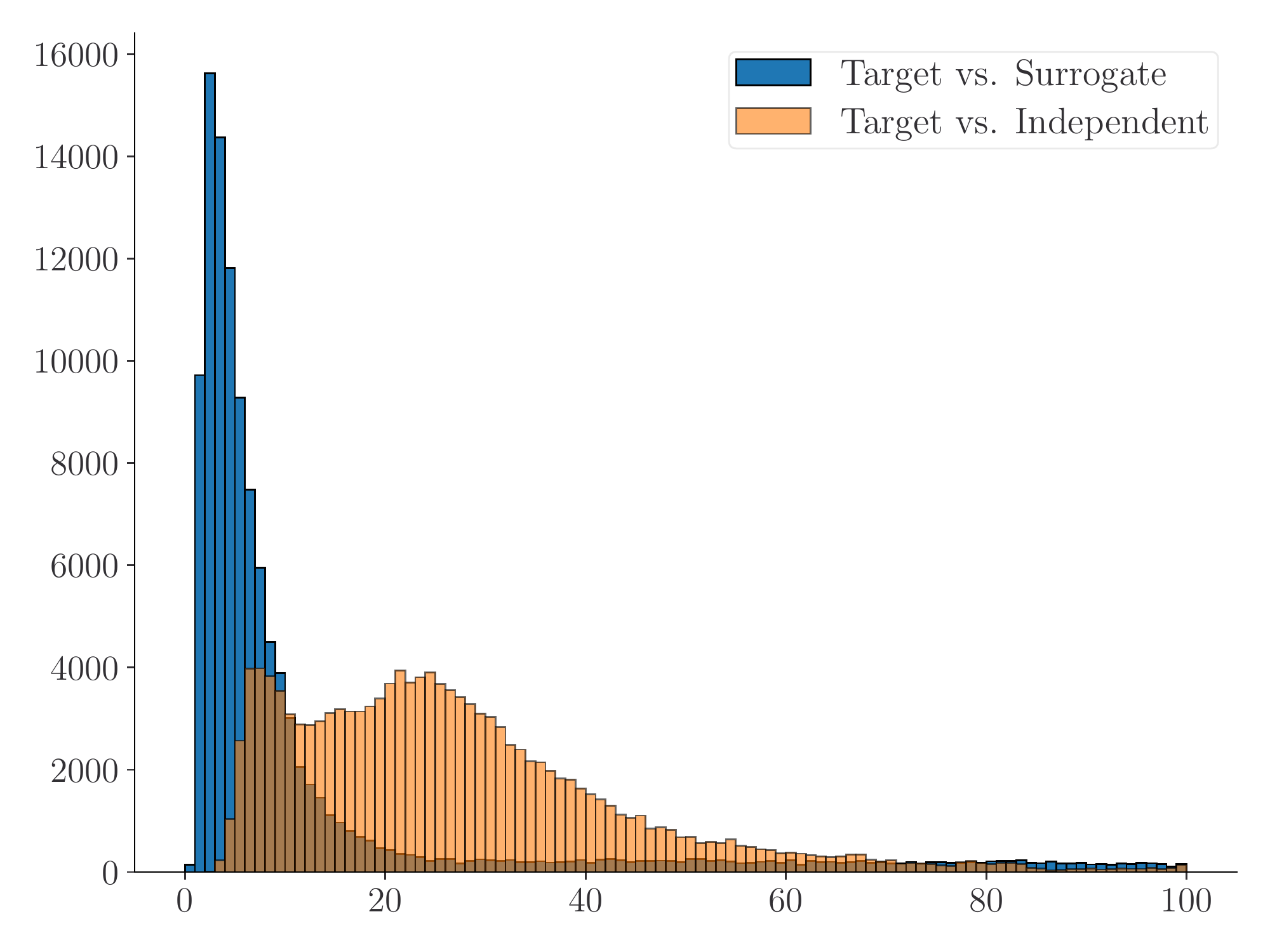}}
    \subfigure[\citseer]{\includegraphics[width=0.23\textwidth]{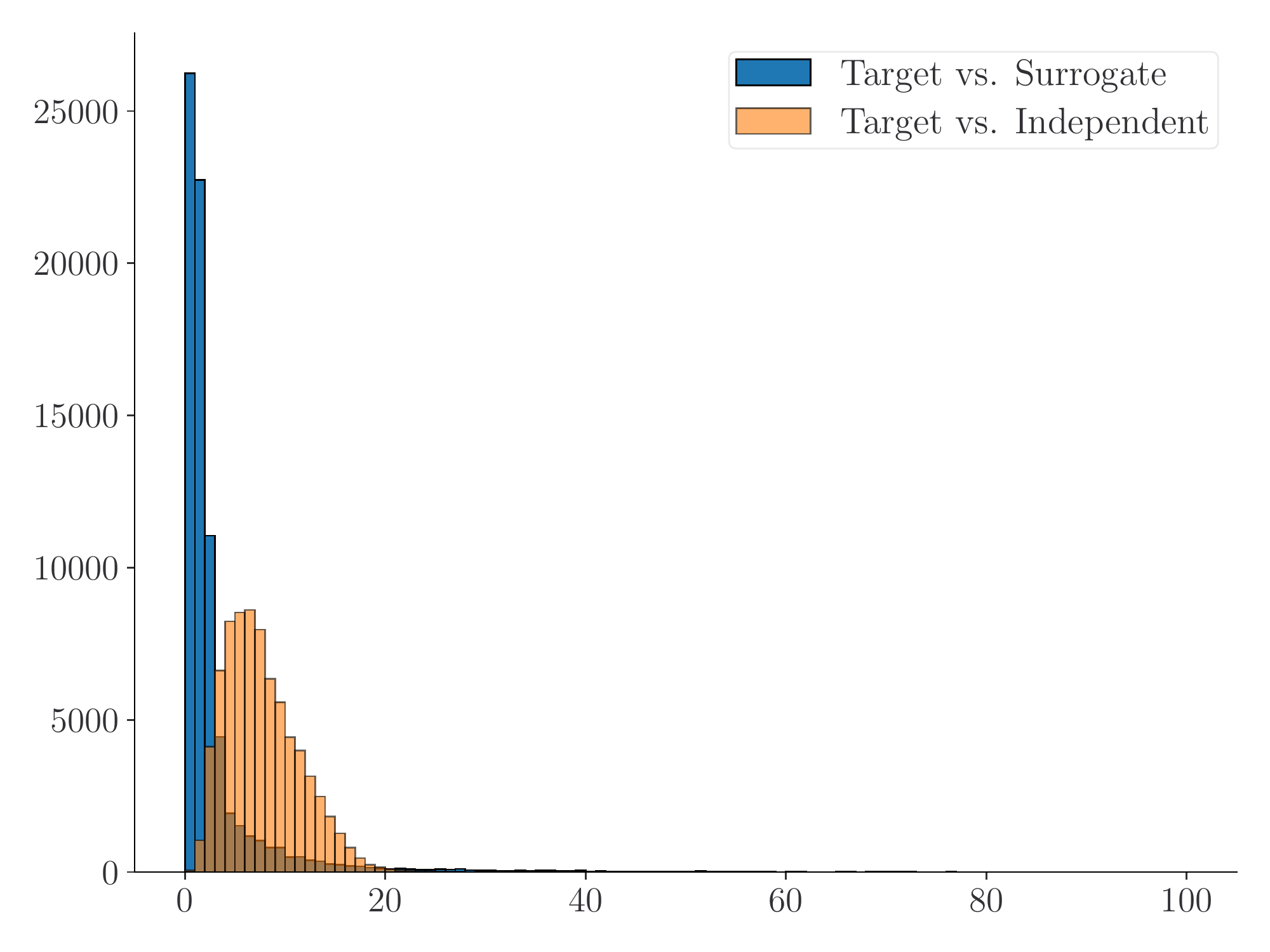}}
    \subfigure[\coauthor]{\includegraphics[width=0.23\textwidth]{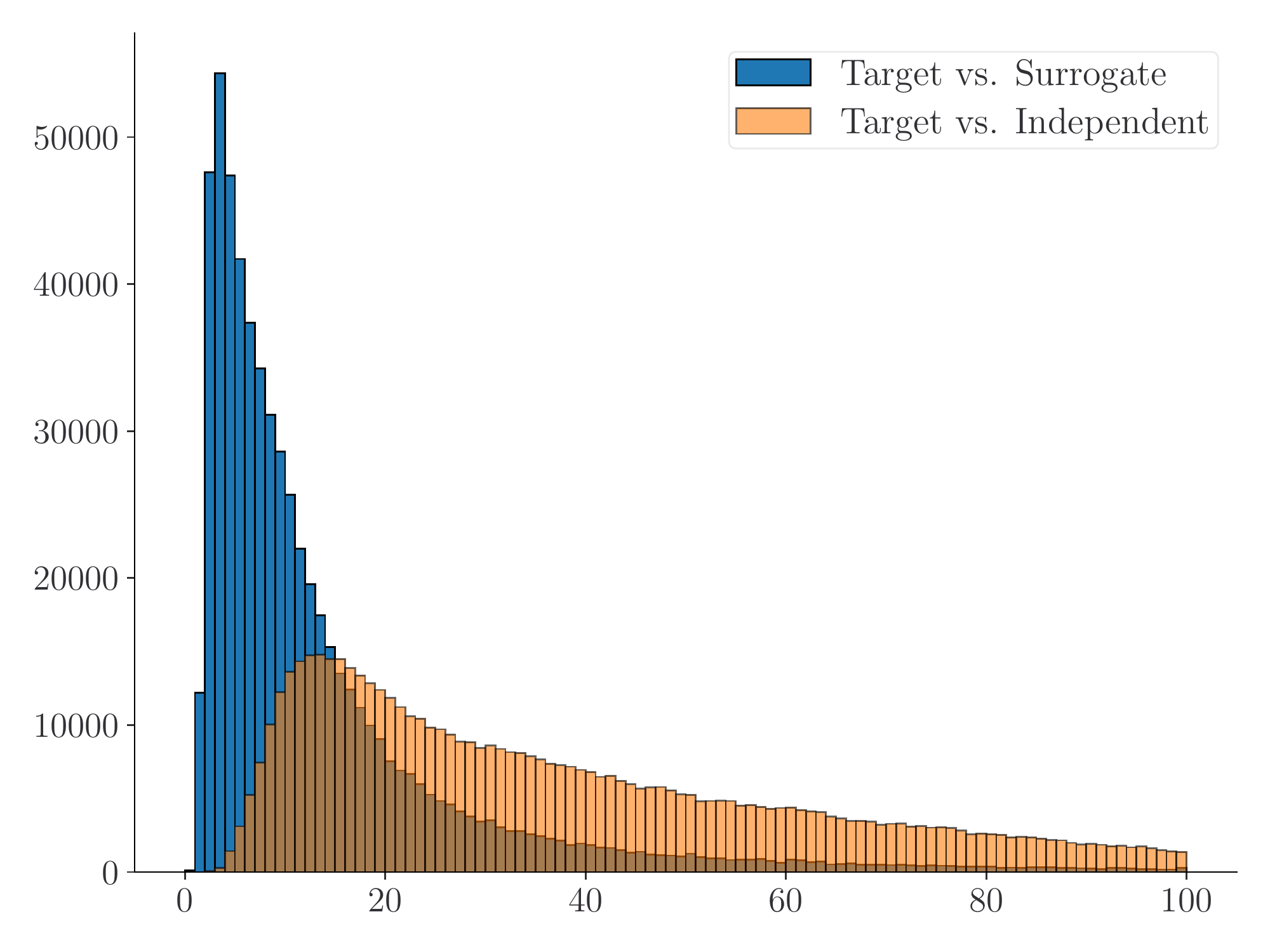}}
    \subfigure[\dblp]{\includegraphics[width=0.23\textwidth]{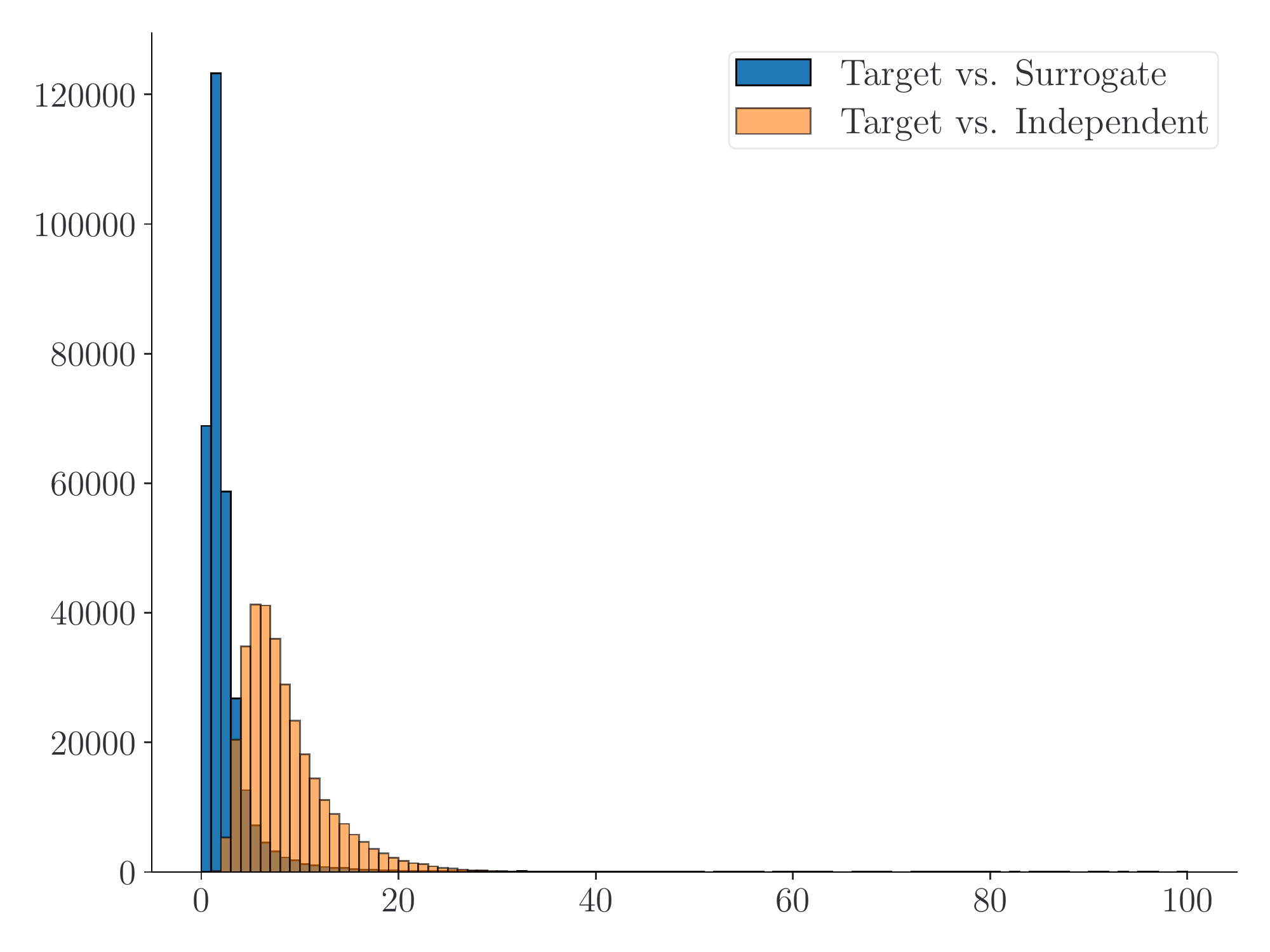}}
    \subfigure[\pubmed]{\includegraphics[width=0.23\textwidth]{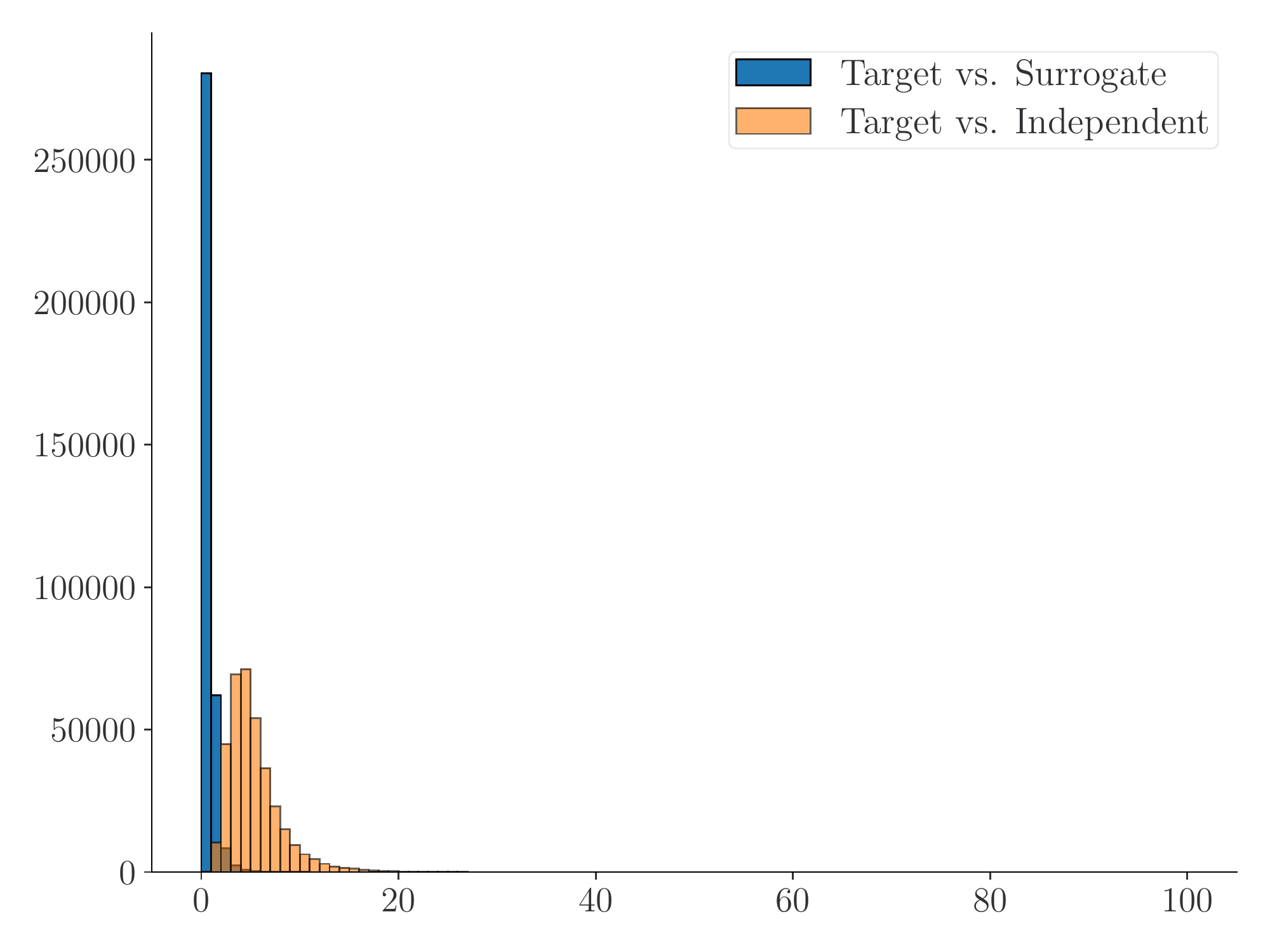}}
    \caption{Histograms of the Euclidean distances between pairs of embeddings from (\target, \surrogate) and (\target, \independent).}
    \label{appendix_fig:distance_plots}
\end{figure}

We calculate the Euclidean distance between pairs of embeddings from (\target, \surrogate) and (\target, \independent) and plot the histogram of the distances in Figure~\ref{appendix_fig:distance_plots}. While the distribution of the distances is different, the distributions overlap significantly. Thus, a simple approach based on distance calculation and hypothesis testing did not work.

\newpage 
\section{Meta-Review}

\subsection{Summary}
In this work the authors present the first fingerprinting mechanism for Graph Neural Networks (GNNs) to facilitate ownership verification.

\subsection{Scientific Contributions}
\begin{itemize}
\item Provides a Valuable Step Forward in an Established Field
\item Addresses a Long-Known Issue
\item Establishes a New Research Direction
\item Creates a New Tool to Enable Future Science
\end{itemize}

\subsection{Reasons for Acceptance}
\begin{enumerate}
\item This paper provides a valuable step forward in an established field. The authors introduce the first fingerprinting mechanism to facilitate ownership verification in graph neural networks (GNNs), thus providing a novel defense against model extraction attacks in this domain.
\item This work addresses a long-known issue. State-of-the-art high-fidelity model stealing attacks are a real threat that has attracted significant research efforts. The mechanism introduced here can be used to defend against these attacks in GNN settings.
\item This work establishes a new research direction. As this is the first mechanism of its kind in GNNs, the authors have identified an area that requires further investigation and exploration. The unique approach introduced here (particularly the use of embeddings) can inspire future work in defending against model extraction and stealing attacks in GNNs.
\item This work provides a new tool capable of verifying ownership of a trained GNN model, which the authors plan to open source and make available for future research.
\end{enumerate}

\subsection{Noteworthy Concerns} 
\begin{enumerate} 
\item The authors show that their approach is robust against an adversary's attempts at detection evasion. They do so by considering simple evasion attacks and model retraining. However, the robustness evaluation would be made considerably stronger by including (a) black box attacks and (b) adaptive attackers that could end-to-end fine tune the model to evade the proposed mechanism with, for example, an additional objective.
\item The authors note that it would not be feasible to empirically compare watermarking/fingerprinting solutions in other domains as it is with GrOVe in the graph domain. While the reviewers believe these baselines would still provide value, we also acknowledge that this is not a significant limitation as watermarking may affect training and other fingerprinting methods are not directly applied to GNNs.
\end{enumerate}

\end{document}